\RequirePackage{fix-cm}
\documentclass[10pt,twocolumn,letterpaper]{article}
\PassOptionsToPackage{table}{xcolor}
\usepackage{times,amssymb}
\usepackage[numbers,sort&compress]{natbib}
\usepackage[font=small,labelfont=bf,labelsep=period]{caption}
\usepackage[T1]{fontenc}
\usepackage{amsmath,amsfonts,bm}

\usepackage[hyphens]{url}
\usepackage{enumitem}
\usepackage{graphicx}
\usepackage{wrapfig}
\usepackage{float}
\usepackage{booktabs}
\usepackage{tabularx}
\usepackage{longtable}
\usepackage{multirow}
\usepackage{xcolor}
\usepackage{array}
\usepackage{placeins}
\usepackage{adjustbox}
\usepackage{fvextra}
\ifdefined\DeclareUnicodeCharacter
\DeclareUnicodeCharacter{2192}{\ensuremath{\rightarrow}}
\fi

\usepackage{microtype}
\usepackage{needspace}
\definecolor{linkblue}{rgb}{0.21,0.49,0.74}
\usepackage[pagebackref,breaklinks,colorlinks,allcolors=linkblue]{hyperref}
\date{}
\title{EOV-Bench: An Evidence-Oriented Benchmark for Long-Horizon Streaming Video Understanding}
\author{Zelin Dai \quad Jianqin Liu \quad Meiling Tao \quad Yang Yang \quad Peng Wang\\
University of Electronic Science and Technology of China}

\begin{document}
\raggedbottom
\setcounter{dbltopnumber}{1}
\setlength{\dbltextfloatsep}{14pt plus 2pt minus 2pt}

\maketitle

\begin{abstract}
Long-horizon streaming video understanding demands fine-grained perception, distant recall, multi-evidence reasoning, and current--history integration under low-latency requirements. Existing benchmarks assess these capabilities unevenly, limiting what high scores reveal about reliable long-horizon understanding. We introduce EOV-Bench, comprising 4,337 open-ended questions over 54.3 hours of video. Its evidence-first construction specifies and combines supporting observations before formulating questions, with human review checking factual correctness, answerability, and the necessity of intended evidence dependencies. Matched historical questions are evaluated in independent causal streaming runs at increasing delays of up to 32 minutes, tracking the continued usability of the same evidence. Evaluation exposes substantial limitations in historical and joint understanding. Recent raw frames favor current perception, while Historical ablations reveal the importance of detailed descriptions, structured memory organization, and effective access to retained information. These findings motivate AsyncMemFlow, a training-free framework that uses asynchronous execution to move fine-grained description and Memory Map construction off the query-response path. Richer memory is prepared in the background and accessed without waiting for ongoing updates. With local 8B models, AsyncMemFlow improves the five-delay mean Historical score by 7.27 percentage points over the strongest baseline in the main comparison, with an average online response latency of 1.31 seconds. Cross-family model and judge evaluations further support the robustness of our results.

\end{abstract}

\section{Introduction}
\label{sec:introduction}

\begin{figure*}[t]
\centering
\includegraphics[width=\linewidth]{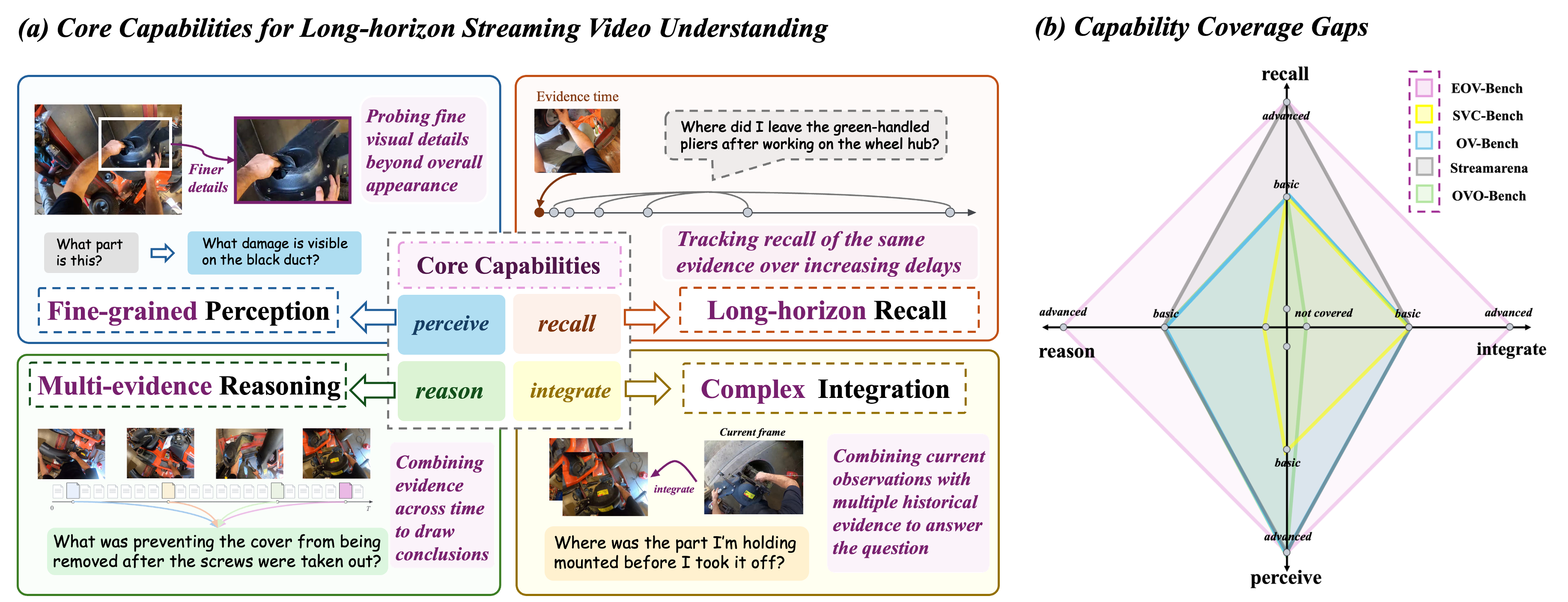}
\setlength{\abovecaptionskip}{3pt}
\setlength{\belowcaptionskip}{0pt}
\caption{\textbf{Challenges and evaluation gaps in long-horizon online video understanding.}
(a) Four key challenges: fine-grained perception, long-horizon recall, multi-evidence reasoning, and current--history integration.
(b) Existing benchmarks provide incomplete coverage of these challenges. The three qualitative levels denote no coverage, basic coverage, and advanced evaluation. Appendix~\ref{app:capability_coverage} details the assessment criteria and coverage analysis; Appendix~\ref{app:ovo_construction_case} provides an evidence-first construction case study.}
\label{fig:reliable_online_evaluation}
\end{figure*}

Long-horizon streaming video understanding requires a system to interpret the present while preserving and reasoning over an increasingly long video history. As video arrives, the system needs to retain useful information without knowing future questions, then answer immediately when they are asked. Analyzing this process through continuous observation, memory updating, and query-time answering (Section~\ref{sec:ovu_formulation}), we identify four essential capabilities: \emph{fine-grained perception} of subtle visual details; \emph{long-horizon recall} of distant observations; \emph{multi-evidence reasoning} that combines independent historical facts to infer relationships no single observation can establish; and \emph{current--history integration}, where answering requires both present and past evidence. These capabilities define the demands of reliable long-horizon video understanding (Figure~\ref{fig:reliable_online_evaluation}).

Existing benchmarks cover these capabilities unevenly and do not consistently assess them in depth (Appendix~\ref{app:capability_coverage}). SimpleStream~\citep{shen2026simplebaselinestreamingvideo} shows that just four recent frames can match or surpass published streaming systems on evaluated subsets. These gaps call for a benchmark with carefully validated, evidence-grounded questions, comprehensive capability assessment, and a strict streaming protocol that reflects the constraints of continuous observation and immediate answering.

We introduce \textbf{EOV-Bench}, a benchmark that makes the evidence requirements of long-horizon streaming questions explicit. We select and combine multi-scale video evidence before formulating questions, specifying the observations and relationships that answering requires. Human reviewers verify both factual support and the necessity of the intended evidence dependencies against the source videos. The resulting 4,337 open-ended questions assess the four capabilities through carefully constructed demands on visual detail, historical information, and cross-time reasoning.

These questions are evaluated under a strict causal streaming protocol: video continues during memory updates, and questions remain unknown until arrival. Matched historical questions are evaluated in independent runs at increasing delays of up to 32 minutes, using duration-eligible subsets at the longest delays. Holding question wording, evidence, and reference answers fixed tracks how the same information remains recoverable as intervening video accumulates.

EOV-Bench reveals limitations that aggregate scores alone obscure. Methods with similar Historical averages exhibit markedly different retention profiles. Recent raw frames excel at current perception but remain weak on historical and joint questions. Controlled Historical ablations show gains from richer descriptions, structured memory organization, and effective access to stored observations. Applying our direct memory access to OASIS~\citep{liang2026oasis} also improves answering without changing its descriptions. Together, these findings identify memory construction and use as complementary targets for improving long-horizon understanding.

Richer memory preparation requires computation, while online answering demands low latency. We therefore introduce \textbf{AsyncMemFlow}, a training-free framework that moves fine-grained description and structured memory organization into an independently configured background pipeline. At query time, the answerer routes each question to recent raw frames or directly reads the latest completed Memory Map, supplemented by available ungrouped descriptions and retained recent frames. Answers do not wait for ongoing updates, allowing more extensive memory preparation to support accurate, responsive online answering.

With local 8B models, AsyncMemFlow improves the five-delay Historical mean by 7.27 points over the strongest baseline in the main comparison, while answering in 1.31 seconds on average and largely preserving raw-frame perception performance. With Qwen-VL-Max descriptions, a 2B answerer surpasses the main baselines at every Historical delay. Cross-family model and judge evaluations, alongside tests on OVO-Bench and StreamArena, support the robustness and broader applicability of the findings. Substantial remaining errors nevertheless highlight the unresolved challenges exposed by EOV-Bench.

\paragraph{Contributions.}
Our contributions are threefold:
(1) EOV-Bench, combining evidence-first question construction with strict causal streaming evaluation to assess four essential capabilities and track historical retention;
(2) systematic diagnoses of distinct retention profiles and the effects of description quality, memory organization, and access on historical answering; and
(3) AsyncMemFlow, which invests background computation in detailed, structured memory to improve memory-dependent answering while maintaining low online response latency.

\section{Related Work}

\subsection{Memory for Streaming Video Understanding}

Streaming video understanding requires preserving useful observations as temporal context grows under limited computational resources~\citep{10654800,zhang2025flashvstream}. Existing methods explore visual memory compression~\citep{10654800,zhang2025flashvstream}, event-level aggregation~\citep{NEURIPS2025_6dd91fec}, compact KV-cache management~\citep{yang2025streammemqueryagnostickvcache,zhang-etal-2026-hermes}, and query-dependent retrieval of historical states~\citep{ICLR2025_67a9b444}. Textual summaries and reasoning traces retain semantic history alongside recent visual observations~\citep{Chatterjee_2025_ICCV,10.1007/978-3-032-37255-0_5,10.1007/978-3-032-37447-9_2}.

Work on textual and hybrid memory explores event hierarchies and entity relations~\citep{liang2026oasis,zhang2026streamarenacontinuousinteractivelonghorizon}, selective semantic retrieval~\citep{fan2026foliofocusedsemanticmemory}, and iterative agentic search~\citep{ICLR2026_ec370e88,wen2026eventmemagenthierarchicaleventcentricmemory,zhang2026streamarenacontinuousinteractivelonghorizon}. Controlled experiments further show benefits from selective memory writing~\citep{fan2026foliofocusedsemanticmemory} and improved memory generation with the answerer held fixed~\citep{ICLR2026_ec370e88}. Meanwhile, SimpleStream shows that additional history can improve recall while degrading current perception, motivating different evidence paths for current and historical questions~\citep{shen2026simplebaselinestreamingvideo}.

Asynchronous execution also enables continuous video processing and memory maintenance alongside online answering, as explored by Flash-VStream, StreamChat, OASIS, and StreamMind~\citep{zhang2025flashvstream,ICLR2025_ad2fa437,liang2026oasis,zhang2026streamarenacontinuousinteractivelonghorizon}. Building on these designs, we investigate how background computation can improve the quality and usability of historical memory. AsyncMemFlow combines detailed descriptions from an independently configured model, a structured Memory Map, and direct memory access to improve historical answering with a compact answerer while maintaining low response latency.

\subsection{Online Video Understanding Benchmarks}

Perception, recall, reasoning, and integration have motivated complementary benchmark designs: timestamped visual questions~\citep{11463959,Huang_2025,11094230}, historical memory and retrieval~\citep{ICLR2025_ad2fa437,ICLR2026_1022661f,zhang2026streamarenacontinuousinteractivelonghorizon}, cross-clip and hierarchical reasoning~\citep{ICLR2025_693ce820,NEURIPS2025_19e4ea30}, and history-dependent spatial understanding and state tracking~\citep{NEURIPS2025_10d3c8a6,li2026ovosbenchhierarchicalbenchmarkstreaming,10.1007/978-3-032-37044-0_29}. These efforts broaden capability coverage, but their evaluation depth and scope remain uneven.

As discussed in Appendix~\ref{app:capability_coverage}, long-delay recall can still depend on a single historical moment~\citep{ICLR2026_1022661f}, while 565 of StreamArena's 877 historical questions use single-point reasoning, limiting the emphasis on combining evidence~\citep{zhang2026streamarenacontinuousinteractivelonghorizon}.  Explicit current--history integration is examined in spatial and counting benchmarks, but these specialized settings leave broader joint reasoning less explored~\citep{NEURIPS2025_10d3c8a6,li2026ovosbenchhierarchicalbenchmarkstreaming,10.1007/978-3-032-37044-0_29}. EOV-Bench addresses this gap through evidence-first construction and strict streaming evaluation, targeting finer perception, longer retention, multi-evidence reasoning, and complex current--history integration.

\section{EOV-Bench}
\label{sec:eovbench}

Deeper evaluation of long-horizon streaming video understanding requires questions that place explicit demands on perception, recall, reasoning, and integration. EOV-Bench makes these demands part of question construction: we first specify and verify the required evidence, then formulate questions around it. This evidence-first, bottom-up process controls what each question tests, while strict streaming evaluation preserves its temporal requirements.

\subsection{A Decoupled Abstraction of Streaming Video Understanding}
\label{sec:ovu_formulation}

For a video stream $\mathcal{X}=(X_1,X_2,\ldots)$, only the observed prefix $X_{\leq t}$ is accessible at time $t$. A model maintains a memory state and answers a separately arriving question $Q_{t_q}$ through
\begin{align}
M_t &= \mathcal{U}_{\theta}(M_{t-1},X_t),
\label{eq:stream_update}\\
A_{t_q} &= \mathcal{F}_{\theta}(Q_{t_q},C_{t_q},M_{t_q}),
\label{eq:stream_response}
\end{align}
where $C_{t_q}$ denotes available observations not yet incorporated into $M_{t_q}$. Memory may contain visual tokens, KV states, text, or structured events. We evaluate answering from information available at question arrival, independently of how a method represents or updates its memory.

This task calls for four complementary capabilities. \textbf{Perception} extracts information from visual observations; \textbf{recall} recovers relevant historical information; \textbf{reasoning} derives conclusions from observed evidence; and \textbf{integration} combines current observations with historical information. These capabilities interact: a question about past events may require both locating the relevant observations and reasoning over them. EOV-Bench therefore characterizes questions by their evidence requirements rather than assigning them to four mutually exclusive capability groups.

\subsection{Evidence Dependencies for Capability Assessment}
\label{sec:evidence_dependency}

To translate these capabilities into testable requirements, we specify the evidence a question will depend on before writing it. Two axes guide this process: the evidence's temporal locality relative to the query and its distribution across the video.

\paragraph{Evidence locality.}
Current questions depend on present observations, Historical questions on past observations, and Joint questions on both. Locality determines whether the required information comes from current perception, historical recall, or their combination. For Joint questions, current and historical evidence are each necessary to determine the answer.

\paragraph{Evidence distribution.}
For Current and Historical questions, Localized questions rely on a single supporting interval, whereas Distributed questions require observations from multiple separated intervals. For Joint questions, this distinction concerns the historical component: Localized links current observations to one past event within a single historical interval, whereas Distributed links them to multiple past events across separated historical intervals. Distributed evidence can support factual retrieval, comparison, or reasoning; its separation alone does not establish a reasoning requirement. Multi-evidence reasoning instead requires a conclusion derived from multiple observations that are individually insufficient.

These axes define the evidence requirements within which we increase assessment depth. Perception targets fine details rather than overall appearance. Recall requires locating and disambiguating historical observations, including recovering and comparing information across moments. Reasoning requires combining necessary premises, while integration requires interpreting current observations using relevant history. Repeating historical questions at increasing delays further measures how well the same information remains recoverable as the stream grows.

\subsection{Evidence-First Construction and Question Setting}
\paragraph{From observations to verified evidence.}
We capture video information at three complementary temporal scales: fine visual details and actions in 5-second clips, local events and state changes in 1-minute clips, and broader progression in 5-minute clips. Together, these observations provide a timestamped evidence pool spanning details and temporal context. For each target question type, we specify evidence locality and distribution, then select observations with the required content and relationships. Cross-scale consistency screening removes unresolved contradictions, and human reviewers verify the selected evidence against the source video. On a selected video, caption comparisons across model families show similar information-category distributions consistent with human captions. We select Qwen-VL-Max for its stronger descriptive coverage; this comparison is detailed in Appendix~\ref{app:construction_details}.

\label{sec:question_protocol}

\begin{figure}[t]
\centering
\includegraphics[width=\linewidth]{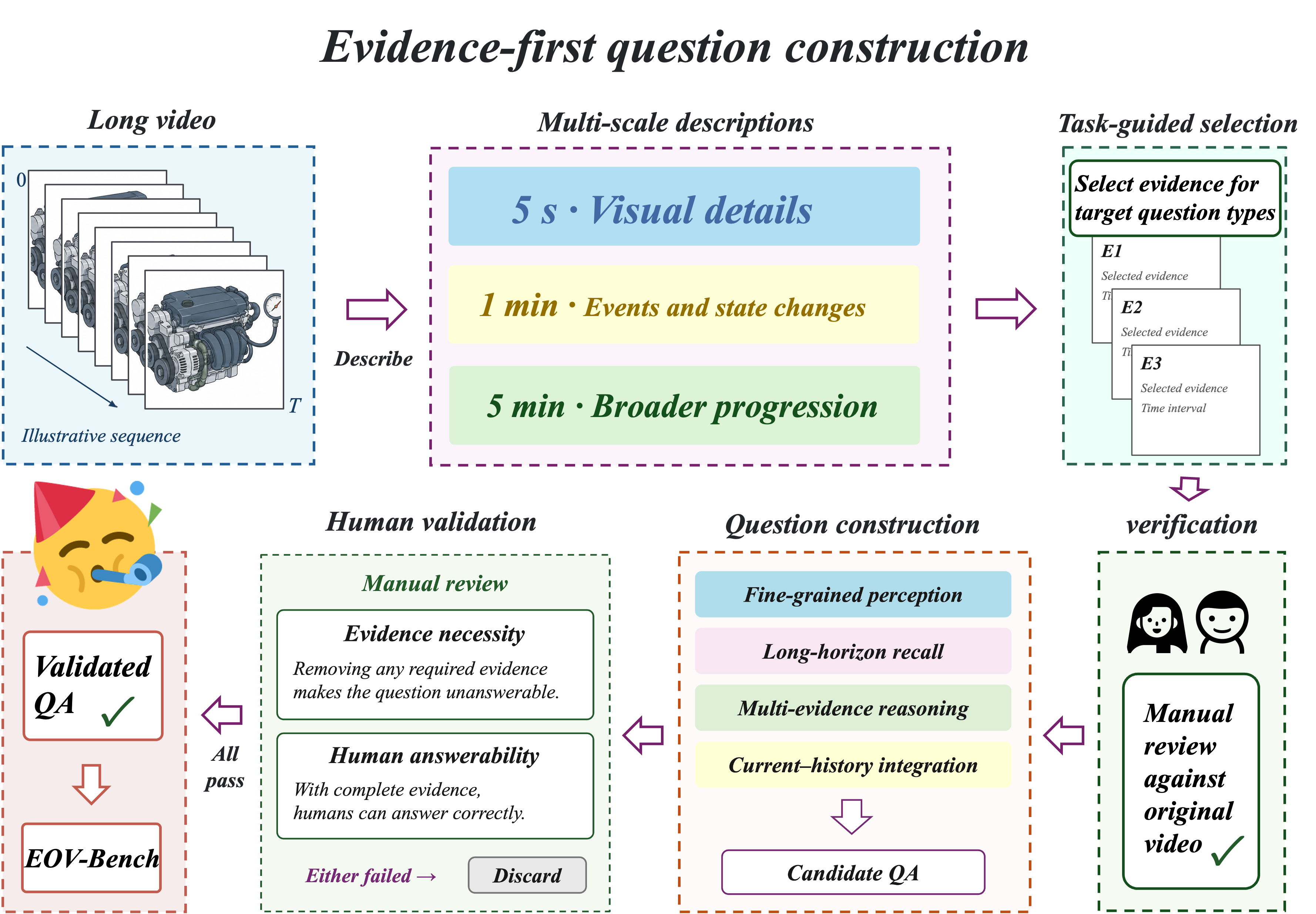}
\caption{\textbf{Evidence-first question construction in EOV-Bench.} }
\label{fig:eovbench_overview}
\end{figure}

\paragraph{Composing evidence into questions.}
Verified observations are used individually or combined to construct questions and reference answers (Figure~\ref{fig:eovbench_overview}). Establishing support before question formulation allows us to select the details to be tested and the observations that answering requires. In particular, cross-moment dependencies are built into the evidence selection rather than inferred from a completed question. A second human review checks answer correctness and whether the intended dependencies are necessary, using the acceptance criteria in Section~\ref{sec:quality_reliability}. Each accepted instance retains its query time, supporting intervals, canonical answer, and accepted alternatives. Full construction details appear in Appendix~\ref{app:construction_details}.

\paragraph{Question families.}
EOV-Bench contains 4,337 open-ended questions over 54.3 hours of video, spanning 49 videos, five domains, and 24 categories. Its 17 question types comprise five Current, seven Historical, and five Joint types. These families organize questions by evidence locality; their capability demands depend additionally on the details and relationships being tested. Appendix~\ref{app:question_examples} presents the complete taxonomy, representative examples, and the correspondence between question types and capabilities.

\subsection{Evaluation Protocol}
\label{sec:evaluation_protocol}

\paragraph{Strict real-time streaming.}
Frames arrive sequentially at 8 FPS without pausing the video clock for memory updates. Questions are released at scheduled times and are answered using information available at their arrival. Future frames are inaccessible, and past observations remain available only if retained by the method.

\paragraph{Matched-question retention.}
To isolate retention of the same information, we evaluate historical questions with unchanged wording, evidence, and reference answers at \textbf{$t_q=b_q+\delta$}, with \textbf{$\delta\in\{30,60,120,240,480,960,1920\}\,\mathrm{s}$},
where $b_q$ is when all supporting evidence has appeared and any question-defined observation window has ended. Each delay uses an independent run, with no questions, answers, or memory states carried over between delays. This prevents earlier repetitions from reinforcing subsequent answers. The first five delays share 1,217 questions; the 16- and 32-minute evaluations use duration-eligible subsets. Comparisons involving these longer delays therefore also reflect cohort changes. Eligibility rules are provided in Appendix~\ref{app:streaming_evaluation}, with question counts in Table~\ref{tab:historical_eligibility}.

\paragraph{Open-ended answer evaluation.}
Models receive no answer options and return at most five words or \texttt{REJECT}. Exact matches to canonical or accepted answers receive full credit. Other responses receive a score in $\{0,0.5,1\}$ from a fixed text-only judge given the question, response, references, and task information. Task-specific rubrics assess the requested facts, relations, directions, or order. Empty, failed, and rejected responses score zero. Appendix~\ref{app:evaluation_prompts} provides scoring rubrics and aggregation details.

\subsection{Data Quality and Evaluation Reliability}
\label{sec:quality_reliability}

\paragraph{Evidence and question acceptance.}
Human reviewers verify selected observations against the source videos for factual accuracy and temporal grounding. Questions are accepted only when their wording is clear, reviewers can recover the reference answer from the designated evidence, and the intended dependencies are necessary. For multi-evidence reasoning, removing any designated supporting observation leaves insufficient evidence to determine the conclusion. For Joint questions, neither current nor historical evidence alone is sufficient. We exclude Historical questions when later footage repeats the answer-bearing evidence or provides alternative evidence that allows later queries to be answered without retaining the designated observations. Questions with reviewer disagreement are discarded. A case study using manually catalogued video information shows that our AI-assisted construction yields broader information coverage, even at matched question counts, and more varied historical assessment than the original benchmark questions (Appendix~\ref{app:ovo_construction_case}). Appendix~\ref{app:question_examples} further analyzes benchmark information coverage and manually annotated video information. We randomly sample 300 model responses from the full evaluation output pool, spanning multiple videos. A reviewer who did not author the corresponding questions independently checks the questions, reference answers, and responses against the source videos. No factual errors are found in the audited questions or reference answers, but two questions admit additional valid answers missing from the references. We add these to the accepted alternatives and recheck all evaluated methods' scores on both questions.

\paragraph{Scorer calibration and validation.}
We refine task-specific scoring prompts against human ratings on a development set randomly sampled in proportion to question-type frequencies and excluded from the main results. After refinement, the Qwen3.5-9B scorer makes no confusions between scores of 0 and 1 on this set; disagreements involving partial credit (0.5 versus 0 or 1) occur in 3\% of cases. We then fix the prompts for all methods. On the same 300 audited responses, comparison with human ratings grounded in the source videos identifies six scoring disagreements (2\%), with automated and human total scores differing by at most one point on the original $\{0,0.5,1\}$ scale. This statistic measures scorer agreement rather than reference-answer coverage. Calibration and audit details appear in Appendix~\ref{app:scorer_validation}. To assess sensitivity to judge family, we rescore the same final model responses using the DeepSeek and GPT APIs under the same scoring criteria. Qwen3.5-9B scores appear in the main results, and DeepSeek and GPT scores in Appendix~\ref{app:broader_ablations}. Key relative method rankings and main conclusions remain consistent across all three judges, supporting robustness to judge choice.

\section{AsyncMemFlow}
\label{sec:asyncmemflow}

Offline video agents demonstrate the value of extended observation, reasoning, and verification~\citep{10.1007/978-3-031-72989-8_4,wang2026thinkverify}, but performing these operations after question arrival increases response latency. Our experiments, consistent with prior observations~\citep{shen2026simplebaselinestreamingvideo,zhang2026streamarenacontinuousinteractivelonghorizon}, reveal complementary requirements: current perception benefits from recent raw frames, whereas historical questions benefit from detailed textual memory. We introduce \textbf{AsyncMemFlow}, a training-free framework that uses asynchronous execution to move fine-grained video description and structured memory organization off the query-response path. This design invests background computation in richer historical memory, enabling more accurate answers with low online response latency.
\begin{figure}[t]
\centering
\includegraphics[width=\linewidth]{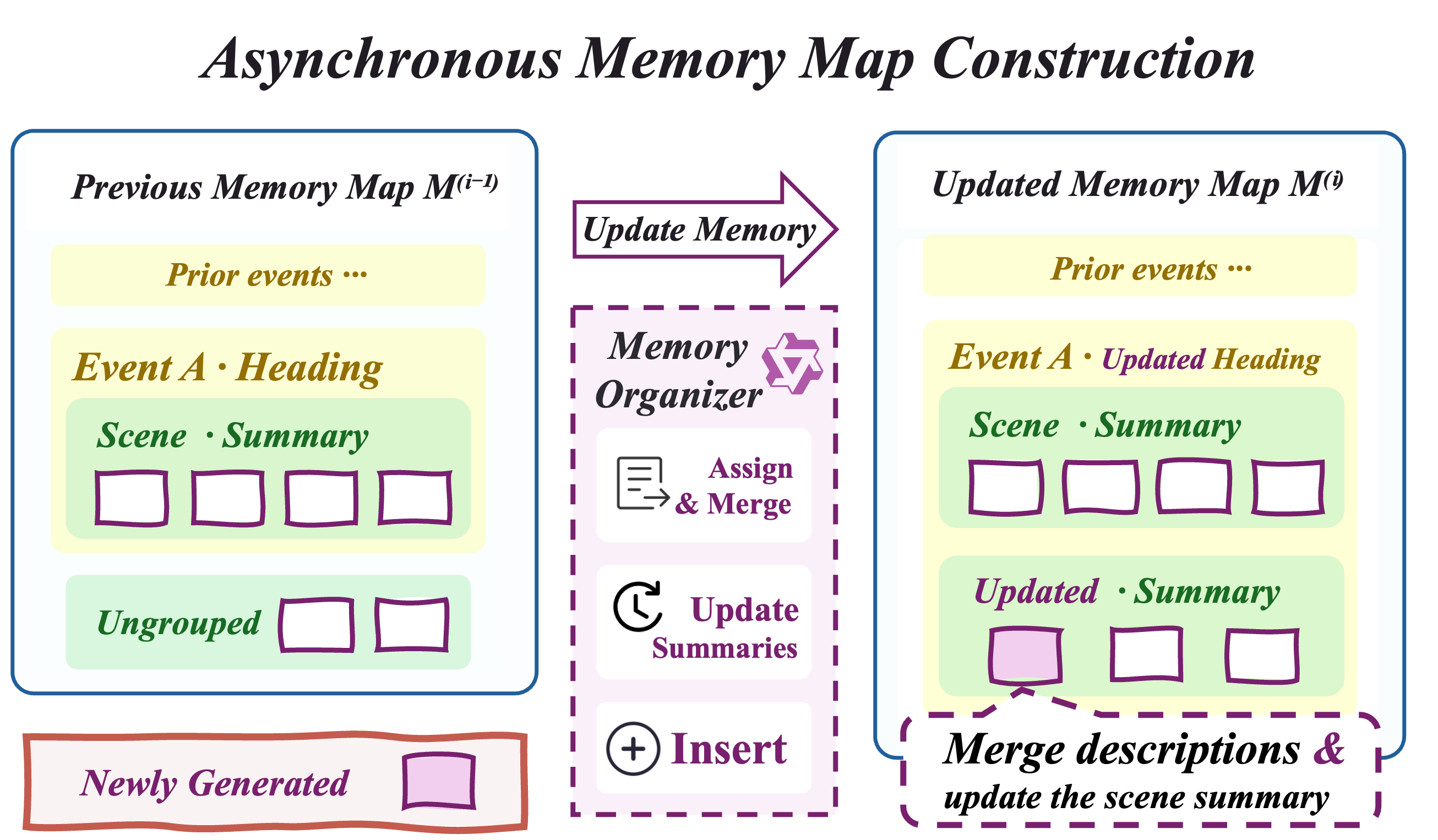}
\caption{\textbf{Asynchronous Memory Map Construction.} }
\label{fig:asyncmemflow}
\end{figure}

\begin{table*}[t]
\centering
\setlength{\abovecaptionskip}{3pt}
\setlength{\belowcaptionskip}{4pt}
\caption{Main results on EOV-Bench (0--100 scores). Bold/underlined scores rank first/second in each column. Historical columns summarize five delays from 30 seconds to eight minutes; Table~\ref{tab:history_results} reports the full seven-delay evaluation.}
\label{tab:current_results}
\label{tab:main_results}
\fontsize{7.5}{9}\selectfont
\setlength{\tabcolsep}{1pt}
\renewcommand{\arraystretch}{1.2}
\begin{adjustbox}{max width=\linewidth}
\begin{tabular}{@{}l*{20}{r}@{}}
\toprule
& \multicolumn{6}{c}{\textbf{Current}}
& \multicolumn{8}{c}{\textbf{Historical}}
& \multicolumn{6}{c}{\textbf{Joint}} \\
\cmidrule(lr){2-7}\cmidrule(lr){8-15}\cmidrule(l){16-21}
& \multicolumn{5}{c}{Localized} &
& \multicolumn{3}{c}{Localized}
& \multicolumn{4}{c}{Distributed} &
& \multicolumn{3}{c}{Localized}
& \multicolumn{2}{c}{Distributed} & \\
\cmidrule(lr){2-6}\cmidrule(lr){8-10}\cmidrule(lr){11-14}
\cmidrule(lr){16-18}\cmidrule(lr){19-20}
\textbf{Method}
& EAR & ECR & OCR & OR & SRR & \textbf{Avg.}
& ER & ESR & LCR & MER & EC & CEC & TRR & \textbf{Avg.}
& CRR & SCR & CHC & MCG & SER & \textbf{Avg.} \\
\midrule

Raw Frames (Qwen3-VL-8B)
& \textbf{83.02} & \textbf{74.60} & \textbf{81.10} & \textbf{70.32} & \textbf{68.15} & \textbf{75.44}
& 1.32 & 2.20 & 1.72 & 7.59 & 0.83 & 1.11 & 22.75 & 5.36
& 10.22 & 4.60 & 5.17 & 11.76 & 6.45 & 7.64 \\
Raw Frames (LLaVA-OneVision)
& 73.92 & 56.51 & 69.23 & 55.36 & 56.60 & 62.32
& 5.38 & 10.50 & 2.50 & 7.04 & 12.50 & 3.52 & 7.50 & 6.99
& 14.52 & 8.64 & 8.19 & 8.82 & 0.00 & 8.03 \\
\midrule
ReKV~\citep{ICLR2025_67a9b444}
& 39.03 & 37.01 & 29.34 & 29.66 & 41.92 & 35.39
& 10.62 & 25.37 & 4.53 & 20.56 & 28.33 & 14.07 & 15.68 & 17.02
& 14.84 & 5.99 & 10.18 & 5.88 & 0.00 & 7.38\textsuperscript{*} \\
HERMES~\citep{zhang-etal-2026-hermes} (Q)
& 67.26 & \underline{70.10} & 69.83 & 61.07 & 64.22 & 66.50
& 20.29 & \underline{36.17} & 17.81 & 32.59 & 46.67 & \underline{23.15} & \underline{58.52} & \underline{33.60}
& \underline{39.25} & 10.66 & 17.24 & 23.53 & 16.13 & 21.36 \\
HERMES~\citep{zhang-etal-2026-hermes} (L)
& 65.54 & 58.58 & 57.01 & 49.30 & 50.91 & 56.27
& 10.74 & 22.23 & 8.28 & 19.81 & 31.67 & 8.33 & 17.63 & 16.95
& 24.73 & 6.80 & 10.78 & 26.47 & 0.00 & 13.76 \\
Flash-VStream~\citep{zhang2025flashvstream} (Q)
& 61.97 & 55.14 & 44.33 & 45.00 & 64.44 & 54.18
& 5.58 & 13.44 & 2.34 & 22.59 & 24.17 & 6.48 & 28.47 & 14.73
& 7.53 & 0.74 & 4.31 & 11.76 & 0.00 & 4.87 \\
OASIS~\citep{liang2026oasis}
& \underline{74.51} & 69.71 & \underline{76.88} & \underline{69.91} & 65.42 & 71.29
& 17.97 & 28.34 & \underline{24.38} & 36.11 & 33.33 & 21.67 & 40.38 & 28.88
& 38.71 & \textbf{43.38} & \underline{41.38} & \textbf{35.29} & \underline{20.97} & \underline{35.95} \\

StreamForest~\citep{NEURIPS2025_6dd91fec}
& 63.98 & 60.77 & 57.33 & 47.44 & 51.12 & 56.13
& 18.75 & 31.13 & 15.31 & 21.11 & \underline{51.67} & 14.26 & 24.62 & 25.26
& 27.96 & 7.72 & 7.33 & 11.76 & 6.45 & 12.24 \\
VST~\citep{10.1007/978-3-032-37255-0_5}
& 67.42 & 69.53 & 69.42 & 52.41 & 51.39 & 62.03\textsuperscript{*}
& \underline{21.06} & 29.57 & 20.59 & \underline{37.00} & 46.67 & 12.09 & 28.38 & 27.91\textsuperscript{*}
& 34.51 & 7.80 & 8.26 & \underline{30.00} & 5.36 & 17.19\textsuperscript{*} \\
Dispider~\citep{Qian_2025}
& 46.56 & 53.12 & 16.58 & 37.85 & 37.06 & 38.23\textsuperscript{*}
& 15.45 & 30.15 & 22.19 & 21.11 & 40.83 & 7.78 & 18.81 & 22.33
& 19.23 & 11.80 & 19.47 & 0.00 & 9.68 & 12.04\textsuperscript{*} \\
\midrule

AsyncMemFlow \textsuperscript{\dag}
& \textbf{83.02}
& \textbf{74.60}
& \textbf{81.10}
& 69.41
& \underline{67.83}
& \underline{75.19}  & \textbf{32.97} & \textbf{40.68} & \textbf{26.56} & \textbf{58.89} & \textbf{53.33} & \textbf{32.78} & \textbf{62.29} & \textbf{43.93} & \textbf{53.76} & \underline{42.46} & \textbf{42.24} & \textbf{35.29} & \textbf{45.16} & \textbf{43.78} \\

\bottomrule
\end{tabular}
\end{adjustbox}
\par\smallskip
\begin{minipage}{\linewidth}
\fontsize{7.5}{9}\selectfont
\textit{Note:}
Historical scores average five delays (30s--8min) over 1,217 shared questions; see Table~\ref{tab:history_results} for all seven delays. Avg. equally weights the 5/7/5 Current/Historical/Joint subtypes. Question-type abbreviations are defined in Appendix~\ref{app:question_examples}. (Q)/(L) denote Qwen3-VL-8B/LLaVA-OneVision backbones. \textsuperscript{\dag}AsyncMemFlow uses separate local 8B models for description and answering. \textsuperscript{*}Runtime errors/OOM count as zero.
\end{minipage}
\end{table*}

\subsection{Asynchronous Memory Construction}
\label{sec:async_memory_construction}
\label{sec:async_semanticization}

AsyncMemFlow allocates background computation to preserve and organize observations before their relevance to future questions is known.

We describe completed one-minute video blocks using 30 chronological frames sampled at 0.5 FPS. The longer observation window provides context for interpreting actions and state changes, allowing individual details to be described alongside the events in which they occur. The describer is instructed to preserve entities, visual attributes, and their relationships in timestamped records for subsequent memory organization. Description runs asynchronously with an independently configured model, allowing richer historical information to be prepared without requiring online answers to wait for memory updates.

\label{sec:memory_organization}
Accumulated descriptions preserve information but do not explicitly organize relationships across time. We construct a hierarchical \textbf{Memory Map} (Figure~\ref{fig:asyncmemflow}): timestamped descriptions form the underlying records, scene summaries organize related observations, and event headings connect related scenes. Summaries provide context for interpreting historical observations, while underlying descriptions retain the details needed for precise answers.

At each background update, the organizer receives the previous Map and newly completed descriptions, incorporates the new observations, and can revise existing entries in light of additional context. Descriptions awaiting grouping remain accessible as ungrouped entries. The update is
\begin{equation}
M^{(j)}=\mathcal{O}_{\psi}\!\left(M^{(j-1)},\mathcal{D}_j\right),
\label{eq:memory_map_update}
\end{equation}
where $\mathcal{D}_j$ denotes the new descriptions supplied to update $j$, and $\mathcal{O}_{\psi}$ is the memory organizer. The answerer reads the resulting Map directly, locating and connecting observations within its structured context without query-dependent retrieval. When the context budget is reached, the organizer compresses the content associated with the earliest event. This conditional compression was not triggered in the EOV-Bench runs, including its longest video.

At question arrival, the answerer uses the latest completed Map without waiting for ongoing background updates.

\subsection{Evidence-Adaptive Online Answering}
\label{sec:evidence_adaptive_query}

The answer model makes a lightweight, text-only routing decision from the question. It is instructed to select the raw-frame branch only when recent visual observations are sufficient to answer. This branch receives the latest five causal frames sampled at 1 FPS, preserving current visual detail while avoiding historical-context processing. All other questions follow the default memory-based path.

The default path receives the latest completed Map and available ungrouped descriptions, together with retained frames from recent history not yet represented in the available textual memory, sampled at 0.2 FPS. Completed descriptions remain accessible while awaiting organization, allowing the answerer to use them without waiting for update.

All inputs are fixed at question arrival, while background processing continues to prepare subsequent memory updates. Sampling settings and the inputs to the two answering branches appear in Appendix~\ref{app:asyncmemflow_configuration}.

\section{Experiments}
\label{sec:experiments}
\begingroup
\setcounter{topnumber}{1}
\setlength{\textfloatsep}{8pt plus 2pt minus 2pt}
\setlength{\floatsep}{8pt plus 2pt minus 2pt}
\setlength{\intextsep}{8pt plus 2pt minus 2pt}

\noindent\textbf{Experimental Setup.}\quad
We follow the strictly causal protocol in Section~\ref{sec:evaluation_protocol}, without pausing the video clock for memory updates. In the main experiments, AsyncMemFlow uses Qwen3-VL-Instruct~\citep{bai2025qwen3vltechnicalreport} answer models at 2B, 4B, and 8B scales, with descriptions generated by either local Qwen3-VL-8B-Instruct or Qwen-VL-Max. In the Qwen-based experiments, the Memory Map organizer is fixed to local Qwen3-VL-8B-Instruct. Our main same-backbone comparisons use Qwen3-VL-8B-Instruct: HERMES (Q), Flash-VStream (Q), and OASIS share this answering backbone with our main AsyncMemFlow configuration. ReKV and HERMES (L) use LLaVA-OneVision; their results are reported separately from the same-backbone comparisons. To assess robustness across backbone families, we additionally adapt all training-free methods to GLM-4.6V-Flash and InternVL2.5-8B. For AsyncMemFlow, each alternative backbone is used for all three roles---video description, Memory Map organization, and question answering---replacing the Qwen models throughout the pipeline. The main conclusions and key relative performance trends remain consistent under these substitutions, supporting robustness beyond the Qwen family, including when memory descriptions are generated by models from other families. Results are reported in Appendix~\ref{app:broader_ablations}, with detailed adaptation and implementation settings in Appendix~\ref{app:evaluation_configurations}. Scores are reported on a 0--100 scale under the shared partial-credit protocol, with scoring details in Appendix~\ref{app:evaluation_prompts}.

\begin{table*}[t]
\centering
\setlength{\abovecaptionskip}{0pt}
\setlength{\belowcaptionskip}{4pt}
\caption{Historical scores across seven query delays and mean response time (s). Bold/underlined values rank first/second.}
\label{tab:history_results}
\label{tab:response_time}
\fontsize{7.5}{9}\selectfont
\setlength{\tabcolsep}{2pt}
\renewcommand{\arraystretch}{1.15}
\begin{tabularx}{\linewidth}{@{}>{\raggedright\arraybackslash}p{0.21\linewidth}*{7}{>{\centering\arraybackslash}X}|>{\centering\arraybackslash}p{0.165\linewidth}@{}}
\toprule
\textbf{Method}
& \textbf{30s} & \textbf{1min} & \textbf{2min}
& \textbf{4min} & \textbf{8min}
& \textbf{16min} & \textbf{32min} & \textbf{Avg. response time (s)} $\downarrow$ \\
\midrule
Raw Frames (Qwen3-VL-8B)
& 8.46 & 5.79 & 4.42 & 4.39 & 3.75 & 3.83 & 3.71 & \textbf{0.384} \\
ReKV
& 16.91 & 17.27 & 17.04 & 18.40 & 15.50 & 15.35 & 13.82 & 0.879 \\
HERMES (LLaVA-OneVision)
& 24.98 & 20.19 & 15.34 & 13.85 & 10.41 & 13.06 & 11.22 & 0.608 \\
HERMES (Qwen3-VL-8B)
& \underline{42.24} & 36.79 & 31.58 & 30.35 & 27.04 & 26.62 & 22.17 & \underline{0.563} \\
StreamForest
& 29.70 & 23.95 & 25.69 & 23.46 & 23.52 & 24.94 & 24.52 & 1.134 \\
Flash-VStream
& 16.52 & 16.65 & 14.26 & 13.00 & 13.20 & 12.72 & 12.94 & 1.186 \\
VST\textsuperscript{*}
& 36.83 & 29.09 & 26.84 & 25.32 & 21.47 & 23.37 & 16.16 & 1.586 \\
Dispider
& 23.35 & 21.58 & 22.89 & 21.08 & 22.74 & 21.88 & 20.79 & 1.382 \\
OASIS
& 33.04 & 31.21 & 27.56 & 26.47 & 26.13 & 24.43 & 23.71 & 6.947 \\
OASIS (w/ Our Memory Strategy)\textsuperscript{\dag}
& 41.65 & \underline{39.21} & \underline{37.20} & \underline{33.86} & \underline{31.36} & \underline{34.17} & \underline{30.77} & 7.122 \\
\midrule
AsyncMemFlow 
& \textbf{49.87} & \textbf{45.26} & \textbf{42.04}
& \textbf{41.46} & \textbf{41.02} & \textbf{42.10} & \textbf{41.58} & 1.311 \\

\bottomrule
\end{tabularx}
\par\smallskip
\begin{minipage}{\linewidth}
\fontsize{7.5}{9}\selectfont
\textit{Note:}
Scores macro-average seven Historical subtypes. The first five delays share 1,217 questions; 16/32-min results use eligible subsets of 1,018/706 questions.
\textsuperscript{*}Runtime errors/OOM score zero and remain in the denominator.
\textsuperscript{\dag} Our direct memory access supplies flattened OASIS descriptions and node summaries to the answerer; see Appendix~\ref{app:oasis_our_version}.
AsyncMemFlow uses an 8B answerer with local 8B descriptions.
\end{minipage}
\end{table*}

\subsection{Main Results}
\label{sec:main_results}
\label{sec:current_perception}
\label{sec:historical_memory}
\label{sec:response_time}

Different capability demands favor different methods (Table~\ref{tab:main_results}). Recent raw frames achieve the highest or tied-highest scores across all five Current subtypes, with a mean of 75.44, but score only 5.36 on Historical and 7.64 on Joint questions. 
Higher Historical scores do not necessarily imply higher Joint scores: HERMES(Q) exceeds OASIS on Historical questions, 33.60 versus 28.88, whereas OASIS leads on Joint questions, 35.95 versus 21.36. These contrasting profiles demonstrate the need to assess current perception, historical understanding, and current--history integration separately: strength in one does not establish comparable performance in another.

Matched-question evaluation distinguishes retention profiles that are obscured by the average Historical score (Table~\ref{tab:history_results}). HERMES(L) and ReKV have nearly identical five-delay means, 16.95 and 17.02, yet their retention profiles differ substantially. HERMES(L) leads at 30 seconds, 24.98 versus 16.91, but falls behind at eight minutes, 10.41 versus 15.50. Holding questions and supporting evidence fixed distinguishes initial answering effectiveness from the continued usability of historical information as intervening video accumulates. Neither the average score nor the short-delay result captures both behaviors.

Historical understanding depends not only on what is retained, but also on how the retained information is accessed. Keeping OASIS's descriptions and node summaries unchanged, our direct-access variant raises its five-delay Historical mean from 28.88 to 36.66 (Table~\ref{tab:history_results}). This gain requires no additional video descriptions, showing that the original memory contains information that the original answering pipeline does not fully exploit. Effective access to stored observations is therefore an important target for improving long-horizon understanding.

\begin{table*}[t]
\centering
\setlength{\abovecaptionskip}{3pt}
\setlength{\belowcaptionskip}{4pt}
\caption{Ablation of the description model, Memory Map organization, and answer model capacity on Historical questions, macro-averaged over seven subtypes. All answer models are Qwen3-VL~\citep{bai2025qwen3vltechnicalreport} models.}
\label{tab:ablation}
\fontsize{7.5}{9}\selectfont
\setlength{\tabcolsep}{3pt}
\begin{tabular*}{\textwidth}{@{\extracolsep{\fill}}ll*{8}{c}@{}}
\toprule
\textbf{Answer model} & \textbf{Description model} & \textbf{Map}
& \textbf{30s} & \textbf{1min} & \textbf{2min}
& \textbf{4min} & \textbf{8min} & \textbf{16min} & \textbf{32min} \\
\midrule
2B & Qwen-VL-Max & $\checkmark$
& 45.95 & 41.28 & 40.09 & 41.02 & 39.61 & 38.47 & 37.64 \\
2B & Qwen-VL-Max & $\times$
& 45.07 & 40.24 & 38.20 & 36.38 & 35.60 & 36.33 & 36.62 \\
4B & Qwen-VL-Max & $\checkmark$
& 52.79 & 49.49 & 45.78 & 44.08 & 46.15 & 46.66 & 47.13 \\
4B & Qwen-VL-Max & $\times$
& 49.83 & 43.22 & 38.37 & 36.10 & 35.07 & 34.87 & 32.12 \\
8B & Qwen-VL-Max & $\checkmark$
& 55.70 & 48.95 & 47.14
& 47.57 & 50.61 & 48.74 & 51.59 \\
8B & Qwen-VL-Max & $\times$
& 52.94 & 47.02 & 41.03 & 41.06 & 42.32 & 42.26 & 43.52 \\
\midrule
8B & Qwen3-VL-8B-Instruct$^{*}$ & $\checkmark$
& 49.87 & 45.26 & 42.04 & 41.46 & 41.02 & 42.10 & 41.58 \\
8B & Qwen3-VL-8B-Instruct$^{*}$ & $\times$
& 44.53 & 35.77 & 30.02 & 27.91 & 26.28 & 27.09 & 27.17 \\
\bottomrule
\end{tabular*}
\par
\begin{minipage}{\textwidth}
\fontsize{7.5}{9}\selectfont
$^{*}$Local Qwen3-VL-8B-Instruct describer, thinking disabled. Map-on/off use identical questions at each delay; cohorts follow Table~\ref{tab:history_results}.
\end{minipage}
\end{table*}

The complete AsyncMemFlow framework improves memory-dependent understanding while largely preserving current perception. With local 8B description and answering models, it achieves a Historical mean of 43.93, exceeding OASIS with our direct memory access by 7.27 points, and a Joint mean of 43.78, exceeding OASIS by 7.83 points. Its Current score of 75.19 remains within 0.25 points of the raw-frame backbone. The local framework answers in 1.311 seconds on average (Table~\ref{tab:description_compute}). Routing accuracy, measured as the proportion of questions correctly classified into Current or default, is 81.41\%, 91.57\%, and 96.61\% with the 2B, 4B, and 8B answerers, respectively. These improvements coexist with substantial remaining errors on Historical and Joint questions, providing further targets for improving long-horizon understanding.

\subsection{Memory Analysis and Ablations}
\label{sec:ablation}
\label{sec:findings}
\label{sec:diagnostic_findings}

\begin{table}[t]
\centering
\setlength{\abovecaptionskip}{3pt}
\setlength{\belowcaptionskip}{4pt}
\caption{Evidence retrieval and Historical performance under different memory access strategies on EOV-Bench.}
\label{tab:memory_access}
\fontsize{7.5}{9}\selectfont
\setlength{\tabcolsep}{4pt}
\renewcommand{\arraystretch}{1.2}

\begin{tabularx}{\linewidth}{@{}>{\raggedright\arraybackslash}Xcc@{}}
\toprule
\textbf{Memory access}
& \textbf{Recall@5 (\%)} $\uparrow$
& \textbf{Historical} $\uparrow$ \\
\midrule

\multicolumn{3}{l}{\textit{AsyncMemFlow}} \\
BM25~\citep{robertson2009bm25}
& 66.92 & 40.31 \\
PPR~\citep{jeh2003personalized}
& 65.55 & 40.52 \\
Word TF-IDF~\citep{salton1988termweighting}
& 65.67 & 40.67 \\
Qwen-guided 
& 72.62 & 42.76\\
Our direct memory access
& -- & 43.93 \\

\midrule
\multicolumn{3}{l}{\textit{OASIS}} \\
Original RAG (Top-5)
& 58.33 & 32.17 \\
Our direct memory access
& -- & 36.66 \\

\bottomrule
\end{tabularx}

\par\smallskip
\fontsize{7.5}{9}\selectfont
\raggedright
\textit{Note:}

Evidence counts as retrieved when a record matches the one-minute video block containing it.
Qwen-guided prompts Qwen to retrieve evidence using a prompt adapted from OASIS~\citep{liang2026oasis}.
OASIS's original RAG retrieval is adjusted to return five blocks for comparison.
Historical scores (0--100) average five delays (30 s--8 min).
Memory content and answer model are fixed within each system.
\end{table}

Broader observation context improves historical answering (Table~\ref{tab:description_compute}). With the local describer, answerer, and Map configuration fixed, extending the description window from 30 to 60 seconds at the same 0.5 FPS raises Historical performance from 34.66 to 43.93. Sampling density is unchanged, but more observations are described together. At the same 60-second window, Qwen-VL-Max further raises the score to 49.99. These results support improving both the temporal context and model capability used to prepare descriptions.

The additional preparation cost translates into much smaller increases in online latency. Completing Map updates with local 30- and 60-second description windows requires 15.52 and 19.72 seconds per video minute, while response latency increases only from 1.274 to 1.311 seconds. Replacing the 60-second local describer with Qwen-VL-Max increases background update time from 19.72 to 37.21 seconds per block, but response latency rises only to 1.496 seconds. The framework thus accommodates more extensive background description while maintaining short online response times.

\begin{table}[t]
\centering
\setlength{\abovecaptionskip}{3pt}
\setlength{\belowcaptionskip}{4pt}
\caption{Memory-update computation, online response latency, and Historical performance on EOV-Bench.}
\label{tab:description_compute}
\fontsize{7.5}{9}\selectfont
\setlength{\tabcolsep}{4pt}
\renewcommand{\arraystretch}{1.2}

\begin{tabularx}{\linewidth}{@{}>{\raggedright\arraybackslash}Xccc@{}}
\toprule
\textbf{Method}
& \shortstack{\textbf{Memory update}\\\textbf{(s/update)} $\downarrow$}
& \shortstack{\textbf{Response}\\\textbf{(s/query)} $\downarrow$}
& \shortstack{\textbf{Hist.}\\$\uparrow$} \\
\midrule

OASIS~\citep{liang2026oasis}
& 7.63 & 6.947 & 28.88 \\

\midrule
\multicolumn{4}{l}{\textbf{\textit{AsyncMemFlow}}} \\

Local 8B, 30s window
& 7.76 & 1.274 & 34.66 \\

Local 8B, 60s window
& 19.72 & 1.311 & 43.93 \\

VL-Max, 60s window
& 37.21 & 1.496 & 49.99 \\

\bottomrule
\end{tabularx}

\par\smallskip
\fontsize{7.5}{9}\selectfont
\raggedright
\textit{Note:}
Memory update measures mean node-generation time for OASIS and mean time per video block to complete a Memory Map update for AsyncMemFlow.
Response measures mean end-to-end online query latency. Both timings are measured in our evaluation.
Hist. averages five delays (30 s--8 min) over 1,217 shared questions.
All AsyncMemFlow configurations use an 8B answer model and Memory Map.
OASIS uses its original configuration.
\end{table}

Memory organization improves the use of accumulated historical information (Table ~\ref{tab:ablation}). Map-off retains all original descriptions and supplies them to the answerer. Nevertheless, with local 8B descriptions, its Historical score falls from 44.53 at 30 seconds to 26.28 at eight minutes as additional video is observed. This decline occurs even though the earlier descriptions remain available: preserving information does not ensure that the answerer can use it effectively within a growing history. Organizing the same descriptions into a Map reduces the decline from 18.25 to 8.85 points, helping sustain performance on the same historical questions as the stream advances. The benefit holds at every evaluated delay across all four configurations. It also persists with stronger Qwen-VL-Max descriptions, where the Map raises the five-delay mean from 44.87 to 49.99.

Query-dependent retrieval does not outperform direct memory access in our evaluation (Table~\ref{tab:memory_access}). Although Qwen-guided retrieval achieves the highest evidence recall among the tested retrievers, its Historical score remains 1.17 points below direct Map access. Our direct memory access also improves OASIS while keeping its descriptions and node summaries fixed. These results support supplying the complete memory context rather than applying top-5 retrieval under the evaluated settings.

\paragraph{Generalization to Other Benchmarks.}
We further evaluate AsyncMemFlow on OVO-Bench~\citep{11094230} and StreamArena~\citep{zhang2026streamarenacontinuousinteractivelonghorizon}. On OVO-Bench, it achieves 80.89, 61.97, and 52.55 on Real-Time Visual Perception, Backward Tracing, and Forward Active Responding, respectively. On StreamArena, it scores 38.02 on real-time perception and 36.63, 34.37, 26.55, and 18.86 on historical-retrospection levels L1--L4. These scores exceed all online methods reported in the StreamArena paper except StreamMind, which uses Qwen3.5-397B-A17B. On L4, where historical evidence precedes the query by more than 30 minutes, AsyncMemFlow even surpasses StreamMind by 1.76 points, supporting the effectiveness of our memory approach.

\section{Conclusion}

We introduced EOV-Bench, combining evidence-first question construction with strict causal streaming evaluation of fine-grained perception, long-horizon recall, multi-evidence reasoning, and current--history integration. Human-verified evidence dependencies clarify question demands, while matched historical queries reveal retention profiles obscured by aggregate scores. Our experiments highlight complementary roles for recent frames and textual history, with historical answering benefiting from detailed descriptions, memory organization, and effective access. These findings motivate AsyncMemFlow, which invests background computation in detailed, structured memory that a compact answerer accesses without waiting for ongoing updates. This design improves historical and joint understanding while maintaining low response latency and largely preserving current perception. Cross-family model and judge evaluations, together with external benchmark results, support the robustness of the main findings. Remaining errors highlight the challenge of keeping historical evidence usable as video accumulates, making EOV-Bench a testbed for further progress in long-horizon streaming understanding.

\endgroup
\setcounter{topnumber}{2}

\Needspace{15\baselineskip}

\clearpage
\onecolumn
\appendix
\section*{Appendix}
\nocite{bai2025qwen25vltechnicalreport}
\section{Benchmark Coverage and Question Quality}
\label{app:benchmark_quality}

\subsection{Capability Coverage and Question Distribution}
\label{app:capability_coverage}

\paragraph{Capability requirements of an individual question.}
Following Sections~\ref{sec:ovu_formulation} and~\ref{sec:evidence_dependency},
we define a capability by the observations a respondent must locate and the
operation needed to answer from them. The query time determines which
observations are current and which are historical. A single question may
exercise several capabilities. Table~\ref{tab:capability_criteria} defines
a basic and a higher-level demand for each capability at the level of one
question.

\begin{table}[H]
\centering
\captionsetup{labelfont=bf}
\caption{\textbf{Four capabilities defined by the evidence and operation a question requires.}
For each capability, the last two columns describe what a respondent must do with the evidence.}
\label{tab:capability_criteria}
\small
\renewcommand{\arraystretch}{1.10}
\setlength{\extrarowheight}{2pt}
\setlength{\tabcolsep}{5pt}
\begin{tabularx}{\linewidth}{@{}>{\raggedright\arraybackslash}p{0.12\linewidth}>{\raggedright\arraybackslash}p{0.18\linewidth}>{\hsize=0.82\hsize\raggedright\arraybackslash}X>{\hsize=1.18\hsize\raggedright\arraybackslash}X@{}}
\toprule
\textbf{Capability} & \textbf{Evidence to locate} & \textbf{Basic demand} & \textbf{Higher-level demand} \\
\midrule
Perception & Visual cues in the current scene. &
Identify a salient object, action, or state visible at query time. &
Resolve a fine detail, or combine current cues to identify it, for example
by locating an object from its position and reading a small marking on it. \\
\addlinespace
Recall & Previously observed facts, possibly from separate moments. &
Recover a specified past fact or a previously tracked state. &
Find the relevant occurrence among similar events in a long history, or
locate facts at several moments and bring them together to summarize or
compare what was observed. \\
\addlinespace
Historical reasoning & Past observations that supply premises for a conclusion. &
Infer a local relationship, for example how an observed action produced
a change within one episode. &
Find observations at different historical moments and combine them to
derive a conclusion. Each supplies a necessary premise and none alone
determines the answer. \\
\addlinespace
Current--history integration & A current observation and relevant past evidence,
each necessary for the answer. &
Match a currently visible entity to its earlier appearance or compare
its present and past states. &
Use history to interpret the current situation, for example by combining
the visible state of a repair with the earlier procedure to infer which
step is still needed. \\
\bottomrule
\end{tabularx}
\end{table}

\paragraph{From capability requirements to question construction.}
We first select and verify the observations that supply the required
information, then write a question that requires the intended operation
(Section~\ref{sec:question_protocol}). For example, given two prices
announced at different moments, asking for both prices requires locating
and reporting two historical facts. Asking for the difference requires
the same recall plus an inference that neither announcement supports
alone. During review of multi-evidence reasoning questions, we check
that removing any designated premise leaves the
conclusion undetermined. For integration, we check that both the current
scene and the historical evidence are needed with the question wording
held fixed (Section~\ref{sec:quality_reliability}).

\paragraph{Coverage across a benchmark.}
For an existing benchmark, we start with its published tasks and released
questions. We identify the evidence and operation they require under the
evaluation protocol, then examine which demands recur for each capability
in Figure~\ref{fig:reliable_online_evaluation}(b).
\textbf{Not covered} means that the examined material does not establish
a recurring assessment of that capability. \textbf{Basic coverage} means
that the recurring demands are basic or that higher-level demands recur
only within a narrow range of operations.
\textbf{Advanced coverage} requires recurring higher-level demands across
different visual distinctions, historical retrieval targets, inference
patterns, or current--history relationships, as appropriate to the
capability. A single question can therefore make a higher-level demand
even when the available evidence supports only Basic coverage for the
benchmark. Recurring demands may occur across different named tasks.

\paragraph{Comparison scope and supporting evidence.}
We compare visual questions about the present or past that can be answered
from evidence available when the question arrives. For OVO-Bench this
includes the five non-predictive Real-Time Visual Perception tasks and the
three Backward Tracing tasks. Future Prediction and Forward Active
Responding, which allows a response to depend on subsequent footage, fall
outside this comparison. We examine OVBench's current- and past-oriented
tasks, all eight SVCBench counting categories, and the visual demands of
StreamArena's Real-Time Multimodal Perception and Historical Retrospection
tasks. OVBench's Future Prediction tasks and StreamArena's tool-use and
proactive tasks fall outside the same scope.
Table~\ref{tab:benchmark_capability_basis} gives the resulting designations
for the four prior benchmarks in Figure~\ref{fig:reliable_online_evaluation}(b).
The later RIVER and StreamingBench analyses concern question composition
and retention protocols.

\begin{table}[H]
\centering
\small
\caption{\textbf{Capability coverage in the examined tasks of prior benchmarks.}
P, R, M, and I denote perception, recall, historical reasoning, and
current--history integration. A and B denote Advanced and Basic.
A dash indicates that the examined tasks do not establish recurring assessment
of the capability. The letters summarize coverage across questions.}
\label{tab:benchmark_capability_basis}
\setlength{\tabcolsep}{5pt}
\renewcommand{\arraystretch}{1.15}
\begin{tabularx}{\linewidth}{@{}lccccX@{}}
\toprule
\textbf{Benchmark} & \textbf{P} & \textbf{R} & \textbf{M} & \textbf{I} & \textbf{Assessment basis} \\
\midrule
OVO-Bench & A & B & -- & -- &
Reading text and distinguishing visual attributes and spatial relations.
Backward Tracing retrieves, verifies, or temporally locates observed facts. \\
OVBench & A & B & B & B &
Fine spatial localization, retrieval of earlier actions and trajectories,
procedural judgments, and comparisons of current and past states. \\
SVCBench & B & B & -- & B &
Visible and accumulated counts, identity tracking, and comparisons of
recent appearances with previously seen entities. \\
StreamArena & A & A & B & B &
Detailed on-screen perception, retrieval across long histories, and
cross-moment operations including price comparison and matching current
entities against earlier mentions. \\
\bottomrule
\end{tabularx}
\end{table}

\paragraph{Reasons for the designations.}
\textbf{OVO-Bench.} Historical assessment centers on recovering or verifying a past observation, often with the relevant event already identified in the question. This supports Basic recall. The examined tasks do not systematically require respondents to combine historical premises to derive a conclusion or to use current and historical evidence that are each indispensable to the answer. We therefore mark historical reasoning and integration with a dash. Perception receives Advanced coverage because fine visual distinctions are assessed across attributes, text, and spatial relations~\citep{11094230}.

\textbf{OVBench.} Historical retrieval is generally bounded by a specified interval or procedural context. Historical inference centers on procedural judgments and direct comparisons, while integration centers on matching an entity or comparing its state across time. These recurring demands support Basic recall, reasoning, and integration. Their scope falls short of the higher-level criteria for disambiguating observations across extended histories, deriving conclusions across varied historical relationships, and using history to interpret a current situation. Fine spatial and visual distinctions support Advanced perception~\citep{Huang_2025}.

\textbf{SVCBench.} Across its eight counting categories, perception, recall, and integration serve a common objective: identifying and tracking the entities or events needed to determine a quantity or its change. This restricted range of answer targets gives Basic coverage in all three capabilities. The tasks do not systematically require historical observations to serve as premises for a further causal, procedural, or relational conclusion, so we mark historical reasoning with a dash~\citep{10.1007/978-3-032-37044-0_29}.

\textbf{StreamArena.} Fine visual distinctions and retrieval of specific observations across extended histories support Advanced perception and recall. Historical reasoning is narrower: the recurring demands center on direct comparisons, with limited coverage of conclusions that require multiple historical premises across varied relationships. Current--history integration likewise centers on matching or comparing present and past observations. The examined tasks do not systematically require respondents to interpret the current situation in light of earlier events or procedures across a range of contexts. We therefore assign Basic coverage to reasoning and integration~\citep{zhang2026streamarenacontinuousinteractivelonghorizon}.

Appendix~\ref{app:question_examples} presents the corresponding capability
analysis and question distributions for EOV-Bench.

\paragraph{Information distribution and temporal coverage.}
We count released question records within each stated task subset and
identify repeated-delay evaluations separately. Our information
classification of OVO-Bench's EPM questions is manually checked and
groups questions by the primary information requested in their answers.
It is distinct from OVO-Bench's official task labels. Other counts use the
stated official categories or explicit question forms. These statistics
describe the concentration of assessment targets; they do not estimate
factual error rates or dependence on a particular annotation model.

Representing a capability does not ensure diverse coverage within it.
Classifying OVO-Bench's EPM questions~\citep{11094230} by the information
requested in their answers, location and object or tool identity together
account for 194 of 297 questions (65.3\%). A similar concentration appears
within OVBench's Action Retrieval category~\citep{Huang_2025}, where
248 of 320 questions (77.5\%) ask for a person's historical
location.\footnote{Released question annotations:
\href{https://github.com/JoeLeelyf/OVO-Bench/blob/f3911d72411e6d19e358e8e044ec733051e86e6d/data/ovo_bench_new.json}{OVO-Bench}
and
\href{https://huggingface.co/datasets/MCG-NJU/OVBench/blob/main/ovbench.json}{OVBench}.}
These distributions place substantial emphasis on recovering where
something was or what it was, while other historical information receives
fewer questions. The concentration can persist within individual videos:
six of the eight questions associated with one RIVER
video~\citep{ICLR2026_1022661f} ask about clothing, footwear, or hair
color.\footnote{Video \texttt{QB7FoIpx8os} in the
\href{https://github.com/OpenGVLab/RIVER/blob/master/annotations/Live-Perception.json\#L6564}{released RIVER questions}.}
Multiple questions can thus repeatedly examine the same information
dimension. The RIVER video is an illustrative case, rather than an estimate
of the prevalence of this pattern across its full video collection.

The range of operations being evaluated is also uneven.
SVCBench~\citep{10.1007/978-3-032-37044-0_29} distinguishes eight counting
categories. Across its 1,000 questions, the entities or events to detect,
track, or deduplicate vary within a common counting objective. Among
StreamingBench's~\citep{11463959} 2,392 current and retrospective visual
questions, its designated Causal Reasoning category contains 128 questions,
only 36 of which carry the official Prior
label.\footnote{\href{https://buaa-colalab.github.io/SVCBench/}{SVCBench taxonomy and statistics};
\href{https://huggingface.co/datasets/mjuicem/StreamingBench/blob/main/StreamingBench/Real_Time_Visual_Understanding.csv}{StreamingBench question annotations}.}
A broad task inventory can therefore coexist with limited representation
of particular historical demands.

Temporal coverage introduces a further distinction between how long
evidence is retained and what recovering it requires. In
StreamArena~\citep{zhang2026streamarenacontinuousinteractivelonghorizon},
explicit quantity or numerical queries constitute 40.3\% of historical
questions, but their proportion changes from 65.2\% in L1 to 33.0\% in
L4.\footnote{\href{https://huggingface.co/datasets/hkuzxc/StreamArena/blob/2960c360deba57b717a17ab66a3ae80f5358a6a3/question.en.jsonl}{StreamArena released questions}.
Explicit numerical forms include ``how many'' and ``what is/was the/that
number''; the requested quantities include counts, frequencies, and
reported numerical values.}
Because these groups contain different questions, cross-horizon results
reflect changes in question composition alongside changes in retention
interval. RIVER~\citep{ICLR2026_1022661f} reuses the same 372
Live-Perception questions at four historical delays. This extends the
retention interval while preserving the underlying question and evidence
requirements. For example, a question about a singer's microphone color
is posed 49.7 minutes after its annotated evidence, with no additional
temporal anchor in its wording. As intervening history accumulates, the
intended historical reference remains
implicit.\footnote{RIVER question \texttt{16Z-XQh9jhk@2}: evidence at
130--150\,s and the very-long-delay query at 3134.41\,s; see the
\href{https://github.com/OpenGVLab/RIVER/blob/master/annotations/Retro-Memory.json\#L30925}{released annotation}.}

Question quality also depends on whether the reference answer fully
addresses the requested information. One StreamArena question asks for
both the number of windows and the software name, whereas its reference
answer provides only ``Focusky.''\footnote{Question
\texttt{XR17BIx7B1w::6} in the
\href{https://huggingface.co/datasets/hkuzxc/StreamArena/blob/2960c360deba57b717a17ab66a3ae80f5358a6a3/question.en.jsonl\#L2005}{released StreamArena annotations}.}
Such cases make the correspondence between the question, supporting
observations, and reference answer important to evaluation. This example
identifies incomplete reference coverage for one question and does not
estimate the benchmark's overall reference-error rate. Together,
these patterns motivate controlling which information is selected for
questioning, how observations are combined, and whether the resulting
answer satisfies the question. Appendix~\ref{app:ovo_construction_case} examines
these choices on a shared video segment.

\subsection{Evidence-First Construction: An OVO-Bench Case Study}
\label{app:ovo_construction_case}

OVO-Bench~\citep{11094230} illustrates how broad task coverage can coexist
with concentrated assessment demands. Among its 631 Backward Tracing
questions, Episodic Memory (EPM) accounts for 297 questions (47.1\%), making
it the largest historical task family. Our classification of these EPM
questions shows that 194 (65.3\%) ask about locations or object identities,
whereas only 18 (6.1\%) ask for functional or contextual judgments. Shape,
material, and other attributes, together with textual information, account
for just six questions (2.0\%). Historical assessment within this family
therefore places substantially greater emphasis on recovering where
something was or what it was than on other information and judgments
supported by the video. These patterns motivate examining coverage below
the level of task labels: which observations are selected, and what
understanding their questions require. These statistics are computed from
the released question annotations.\footnote{\href{https://github.com/JoeLeelyf/OVO-Bench/blob/f3911d72411e6d19e358e8e044ec733051e86e6d/data/ovo_bench_new.json}{OVO-Bench released question annotations}.}

To examine how question construction shapes this coverage, we use an
eight-minute first-person repair video from OVO-Bench.\footnote{The released
video identifier is
\texttt{5e1a8326-b006-423e-8294-0f9559c8eb18.mp4} in the
\texttt{Ego4D/clips} directory. Its eight original question identifiers are
23--26 and 54--57 in the released OVO-Bench annotations.}
We manually review the video segment from beginning to end and catalogue
144 independently verifiable facts, each counted as one information item.
We compare the eight original questions with 27 questions constructed
through our evidence-first process. The manual inventory and both
question sets cover exactly the same video segment. This case examines
both the breadth of content selected for assessment and the depth of
historical understanding required. The two question sets use the same
classification rules: each question is assigned to its primary answer
target in Table~\ref{tab:ovo_case_information} and its primary answering
operation in Table~\ref{tab:ovo_case_demands}.

The original questions concentrate on locations: five of eight questions
ask where an object was stored, observed, or placed. Other information
within the repair sequence---including manipulation methods, component
conditions, and relationships between operations---provides opportunities
for a broader assessment. Starting from this evidence, we construct
questions that address these additional aspects, as summarized in
Table~\ref{tab:ovo_case_information}.

\begingroup
\small
\renewcommand{\arraystretch}{1.08}
\setlength{\extrarowheight}{2pt}
\setlength{\tabcolsep}{5pt}
\setlength{\LTpre}{8pt}
\setlength{\LTpost}{10pt}
\setlength{\LTcapwidth}{\linewidth}
\begin{longtable}{@{}>{\raggedright\arraybackslash}p{0.20\linewidth}>{\raggedright\arraybackslash}p{0.29\linewidth}>{\raggedright\arraybackslash}p{\dimexpr0.51\linewidth-4\tabcolsep\relax}@{}}
\captionsetup{labelfont=bf}
\caption{\textbf{Information queried by the two question sets.} Each question is
assigned to its primary answer target; parentheses indicate question counts.}
\label{tab:ovo_case_information}\\
\toprule
\textbf{Information target} & \textbf{Original questions} & \textbf{Evidence-first questions} \\
\midrule
\endfirsthead
\multicolumn{3}{c}{\tablename\ \thetable\ (continued)} \\
\toprule
\textbf{Information target} & \textbf{Original questions} & \textbf{Evidence-first questions} \\
\midrule
\endhead
\midrule
\endfoot
\bottomrule
\endlastfoot
Locations and support relations & Tool storage, extinguisher location, and discarded grass (5) & Pliers after use, placement of removed components, and wheel support across repair stages (3) \\
Colors & Broom color (1) & Pliers-handle and spray-trigger colors (2) \\
Action occurrence & Whether the operator touched a pipe (1) & --- \\
Quantities and frequencies & Power tools on the workbench (1) & Tool-chest visits, removed components, and grass-discard events (3) \\
Actions and manipulation methods & --- & Tire preparation, wheel rotation, and methods of removing the tag and grass (3) \\
Object and tool identities & --- & Retrieved attachments, the tire-levering tool, and tools used across visits and operations (5) \\
Object states and contents & --- & Tire--rim state, accumulated grass, and contrasting component conditions (3) \\
\noalign{\nobreak}
Materials and substances & --- & Tape on the duct, liquid applied to the tire, and material removed from the housing (2) \\
\noalign{\nobreak}
Local action--effect relationships & --- & Actions releasing the tag and exposing the grass-filled opening (2) \\
\noalign{\nobreak}
Temporal order & --- & Ordering of preparation, tool handling, component placement, and cleanup actions (4) \\
\midrule
\noalign{\nobreak}
\textbf{Total questions} & \textbf{8} & \textbf{27} \\
\noalign{\nobreak}
\textbf{Information categories covered} & \textbf{4} & \textbf{9} \\
\end{longtable}
\endgroup

Coverage expands from four to nine information categories, while the
largest category decreases from 62.5\% to 18.5\% of the questions. This
broader distribution persists at the same question count: uniformly
selecting eight questions from our set yields an average coverage of 5.85
categories, compared with four in the original set. Evidence-first
construction therefore produces a more varied selection of information
targets, beyond the increase in question quantity.

The matched-count comparison is the exact expectation under uniform
sampling of eight distinct questions without replacement. If $n_c$ is
the number of questions in category $c$ in the 27-question set, the expected
number of represented categories is
\[
\mathrm{E}[K_8]=\sum_{c:n_c>0}
\left(1-\frac{\binom{27-n_c}{8}}{\binom{27}{8}}\right)\approx5.8484.
\]
Here the nine nonempty category counts are $3,2,3,3,5,3,2,2,4$.
This measure concerns the diversity of information categories, rather than
the fraction of all 144 inventory items covered by the questions.

The difference also concerns how historical information is examined.
The original questions predominantly request individual facts. Our
questions additionally require distinguishing separate occurrences,
comparing observations, aggregating events, and recovering their order
(Table~\ref{tab:ovo_case_demands}).

\begin{table}[H]
\centering
\captionsetup{labelfont=bf}
\caption{\textbf{Assessment depth in the repair-video case.} Questions are grouped
by their primary answering operation.}
\label{tab:ovo_case_demands}
\small
\renewcommand{\arraystretch}{1.08}
\setlength{\extrarowheight}{2pt}
\captionsetup{skip=5pt,width=0.84\linewidth}
\setlength{\tabcolsep}{7pt}
\begin{tabularx}{0.84\linewidth}{@{}>{\raggedright\arraybackslash}X>{\centering\arraybackslash}p{0.09\linewidth}>{\centering\arraybackslash}p{0.14\linewidth}@{}}
\toprule
\textbf{Assessment demand} & \textbf{Original} & \textbf{Evidence-first} \\
\midrule
Single-target factual recall & 7 & 10 \\
Counting objects within a local scene & 1 & 0 \\
Retrieving facts from separate events & 0 & 4 \\
Comparing observations across events & 0 & 4 \\
Aggregating event occurrences & 0 & 3 \\
Recovering temporal order & 0 & 4 \\
Identifying an action responsible for a local effect & 0 & 2 \\
\midrule
\textbf{Total} & \textbf{8} & \textbf{27} \\
\bottomrule
\end{tabularx}
\end{table}

For example, the original set asks where the pliers were before being
picked up. Our set also asks which items were retrieved during two separate
tool-chest visits, in order, requiring the two visits to be distinguished
and both retrievals recovered. Another question compares the visible
problems of two removed components: a split in the duct and grass
accumulation inside the cover. These questions connect specific historical
observations rather than repeatedly requesting isolated locations.
Overall, 15 of the 27 questions explicitly involve cross-event retrieval,
comparison, aggregation, or temporal ordering.

This case illustrates how evidence selection shapes question quality.
Identifying informative observations before formulating questions exposes
both underrepresented content and meaningful relationships across the
video. In this case, the resulting question set examines a broader range
of information and places more varied demands on historical understanding.
The comparison controls the source video and, for category coverage, the
number of questions. It does not establish that every video would show
the same gain or that AI-assisted construction is free of selection bias.
\FloatBarrier
\FloatBarrier
\clearpage
\section{Question Types and Capability Coverage}
\label{app:question_examples}
\label{app:taxonomy_scope}

\subsection{Question Types and Evidence Requirements}
\label{app:capability_requirements}
EOV-Bench contains 4,337 distinct questions across 17 types. Table~\ref{tab:question_setting} identifies the evidence and operation each type requires, following the four capability definitions in Appendix~\ref{app:capability_coverage}. Current questions draw on the scene at query time, Historical questions draw on earlier observations, and Joint questions require both. For a Joint question, \emph{Localized} denotes a single historical interval and \emph{Distributed} denotes multiple separated historical intervals. Several capabilities can contribute to one answer. Recovering facts from different moments exercises recall. Deriving a conclusion that each fact alone leaves undetermined also exercises reasoning.

Of the full Historical inventory, 1,217 questions form the common set evaluated at delays from 30 seconds to eight minutes (Appendix~\ref{app:streaming_evaluation}). Each contributes one question to the inventory regardless of how many times it is asked. Tracing tasks are excluded from the 17 evaluated types.

\begingroup
\small
\setlength{\tabcolsep}{4pt}
\renewcommand{\arraystretch}{1.08}
\setlength{\LTpre}{5pt}
\setlength{\LTpost}{4pt}
\begin{longtable}{@{}>{\raggedright\arraybackslash}p{0.20\linewidth}
                      >{\raggedright\arraybackslash}p{0.31\linewidth}
                      >{\raggedright\arraybackslash}p{\dimexpr0.41\linewidth-6\tabcolsep\relax}
                      >{\raggedleft\arraybackslash}p{0.08\linewidth}@{}}
\caption{\textbf{The 17 evaluated question types.} Rows give the evidence scope, principal capability demands and an illustrative question. Counts refer to distinct questions in the full benchmark.}\label{tab:question_setting}\\
\toprule
\textbf{Question type} & \textbf{Capability and answering operation} & \textbf{Example} & \textbf{Count} \\
\midrule
\endfirsthead
\multicolumn{4}{@{}l}{\emph{Table~\thetable\ continued}}\\
\toprule
\textbf{Question type} & \textbf{Capability and answering operation} & \textbf{Example} & \textbf{Count} \\
\midrule
\endhead
\midrule
\endfoot
\bottomrule
\endlastfoot
\multicolumn{4}{@{}l}{\textbf{Current} \emph{(localized evidence)}}\\*
Optical Character Recognition (\textbf{OCR}) & \emph{Perception.} Locate and read the requested text. & What does the smaller line beneath the station name on the carriage sign say? & 410 \\
\addlinespace[2pt]
Entity-Attribute Recognition (\textbf{EAR}) & \emph{Perception.} Distinguish the requested attribute of a specific entity. & What color is the narrow stripe on the insulation of the wire being soldered? & 398 \\
\addlinespace[2pt]
Object Recognition (\textbf{OR}) & \emph{Perception.} Identify the object specified by its visible context. & What small object is attached to the clip on the coach's collar? & 294 \\
\addlinespace[2pt]
Spatial Relation Recognition (\textbf{SRR}) & \emph{Perception.} Locate two visual features and report their spatial relation. & Where is the thermos cap relative to the junction of the vertical and horizontal foam blocks? & 447 \\
\addlinespace[2pt]
Entity Counting (\textbf{ECR}) & \emph{Perception.} Count visible entities satisfying the stated conditions. & How many separate foam blocks support the left end of the horizontal bar? & 339 \\
\addlinespace[2pt]
\multicolumn{4}{@{}l}{\textbf{Historical} \emph{(localized evidence)}}\\*
Event Recall (\textbf{ER}) & \emph{Recall.} Locate a past occurrence and recover the requested action. & During the argument outside the café, what did the light-shirted man do immediately before leaving? & 255 \\
\addlinespace[2pt]
Entity-State Recall (\textbf{ESR}) & \emph{Recall.} Locate an earlier appearance and recover the entity's state. & During the lecturer's first bicycle-wheel demonstration, how was the axle oriented when he first raised the wheel? & 285 \\
\addlinespace[2pt]
Localized Causal Reasoning (\textbf{LCR}) & \emph{Recall and reasoning.} Link an action to its observed effect within a past episode. & What action caused the objects to slide off the desk during the office confrontation? & 266 \\
\addlinespace[2pt]
\multicolumn{4}{@{}l}{\textbf{Historical} \emph{(distributed evidence)}}\\*
Multi-Event Reasoning (\textbf{MER}) & \emph{Recall and reasoning.} Combine facts from separate past events to infer a conclusion that requires each fact. & At the gas-station stop, how much additional fuel was needed to finish the planned route, given the fuel remaining, the remaining distance, and the stated fuel-consumption rate? & 256 \\
\addlinespace[2pt]
Event Counting (\textbf{EC}) & \emph{Recall.} Recover and count occurrences within the specified period. & During the pursuit leading to his first arrival at Kingston station, how many times did the man switch means of transport? & 257 \\
\addlinespace[2pt]
Cross-Event Comparison (\textbf{CEC}) & \emph{Recall.} Retrieve the properties of separate past events and compare them. & How did the damage visible at the end of the kitchen-fire sequence differ from that at the end of the barn-fire sequence? & 277 \\
\addlinespace[2pt]
Temporal Relation Reasoning (\textbf{TRR}) & \emph{Recall, with reasoning for inferred relations.} Recover the order or overlap of past events. & Had the lecturer finished his first bicycle-wheel demonstration when he began the book demonstration? & 290 \\
\addlinespace[2pt]
\multicolumn{4}{@{}l}{\textbf{Joint} \emph{(localized historical evidence)}}\\*
Cross-Temporal Reference Resolution (\textbf{CRR}) & \emph{Integration.} Select a referent in the current scene and recover its identity or association from history. & Who is inside the sack the porter is lifting now? & 93 \\
\addlinespace[2pt]
State-Change Reasoning (\textbf{SCR}) & \emph{Integration.} Identify the earlier and current states of the same entity and describe the change. & How has the fastening on the freight-car door shown now changed since the man last secured it? & 136 \\
\addlinespace[2pt]
Current--History Comparison (\textbf{CHC}) & \emph{Integration.} Compare a detail visible now with its counterpart in an earlier event. & How does the cleaning tool being used on this component now differ from the one used when it was first removed? & 116 \\
\addlinespace[2pt]
\multicolumn{4}{@{}l}{\textbf{Joint} \emph{(distributed historical evidence)}}\\*
Multi-Event Context Grounding (\textbf{MCG}) & \emph{Integration and reasoning.} Use earlier events to infer what the current situation requires. & Which cleaning step is still needed for the area being inspected now, based on the earlier demonstrated procedure? & 119 \\
\addlinespace[2pt]
State-Evolution Reasoning (\textbf{SER}) & \emph{Integration.} Place the current state at the end of a sequence of earlier changes. & What sequence of transport changes connects the man's departure from the station to the vehicle he has just boarded? & 99 \\
\end{longtable}
\endgroup

\subsection{Information Assessed by the Final Questions}
\label{app:information_coverage}
Table~\ref{tab:eov_information_targets} groups the 4,337 final questions by the information their answers require. Each question counts once. For questions spanning multiple information categories, human annotators assigned each question to the category they judged most central to answering it. Human video annotations record facts such as object attributes, actions, and state changes. Appendix~\ref{app:ovo_construction_case} uses a human inventory to compare what the original OVO-Bench questions and our questions ask about on the same video. Appendix~\ref{app:caption_family_comparison} reports the fraction of human-annotated facts captured by each description model on the sampled footage.

\begin{table}[H]
\centering
\small
\setlength{\extrarowheight}{2pt}
\captionsetup{labelfont=bf,skip=5pt}
\caption{\textbf{Primary information targets across EOV-Bench.} Each cell gives the number of questions and the percentage within its column.}
\label{tab:eov_information_targets}
\setlength{\tabcolsep}{4pt}
\renewcommand{\arraystretch}{1.08}
\begin{tabular}{@{}lrrrr@{}}
\toprule
\textbf{Information target} & \textbf{Current} & \textbf{Historical} & \textbf{Joint} & \textbf{Overall} \\
\midrule
Locations and spatial relations & 447 (23.7\%) & 129 (6.8\%) & 35 (6.2\%) & 611 (14.1\%) \\
Entity, object, and tool identities & 294 (15.6\%) & 82 (4.3\%) & 120 (21.3\%) & 496 (11.4\%) \\
Colors and other visual attributes & 112 (5.9\%) & 110 (5.8\%) & 40 (7.1\%) & 262 (6.0\%) \\
Visible text & 410 (21.7\%) & 23 (1.2\%) & 7 (1.2\%) & 440 (10.1\%) \\
Action occurrence & 246 (13.0\%) & 71 (3.8\%) & 0 (0.0\%) & 317 (7.3\%) \\
Actions and manipulation methods & 0 (0.0\%) & 211 (11.2\%) & 90 (16.0\%) & 301 (6.9\%) \\
Object states and contents & 30 (1.6\%) & 151 (8.0\%) & 143 (25.4\%) & 324 (7.5\%) \\
Materials and substances & 10 (0.5\%) & 47 (2.5\%) & 16 (2.8\%) & 73 (1.7\%) \\
Quantities and frequencies & 339 (18.0\%) & 337 (17.9\%) & 17 (3.0\%) & 693 (16.0\%) \\
Temporal relationships & 0 (0.0\%) & 290 (15.4\%) & 61 (10.8\%) & 351 (8.1\%) \\
Causal and other derived relationships & 0 (0.0\%) & 435 (23.1\%) & 34 (6.0\%) & 469 (10.8\%) \\
\midrule
\textbf{Total} & \textbf{1,888 (100.0\%)} & \textbf{1,886 (100.0\%)} & \textbf{563 (100.0\%)} & \textbf{4,337 (100.0\%)} \\
\bottomrule
\end{tabular}
\end{table}
\FloatBarrier
\clearpage
\section{Benchmark Construction Details}
\label{app:construction_details}

\subsection{Choice of Description Model}
\label{app:caption_family_comparison}

We randomly selected one video from each of the five domains and sampled
noncontiguous intervals totaling five minutes per video. Human annotators
and all three models described the same footage. We grouped the recorded
facts using the information categories in Table~\ref{tab:ovo_case_information}.
For the video discussed in the main text, the three models and the human
annotators showed similar distributions across these categories.

We measured coverage as the fraction of human-annotated facts captured by
a model's descriptions. Coverage was 87.37\% for Qwen-VL-Max, 77.01\% for
Qwen-VL-Plus, and 68.89\% for InternVL2.5. We selected Qwen-VL-Max to
describe the benchmark videos because it recovered the largest share of
the human-annotated facts.

\subsection{Benchmark Construction Procedure}
\label{app:construction_procedure}

\paragraph{Step 1. Describe the video at three temporal scales.}
Qwen-VL-Max describes the source video at three temporal scales
(Table~\ref{tab:construction_information_scales}). Each description is
generated directly from the corresponding video clip and retains the
video identifier and clip interval. Five-second clips capture details
that a longer account might omit, such as a marking on an object, the
location of visible text, or a brief change in state. Unclear identities
and unreadable text can be marked as uncertain. One-minute clips place
actions and changes to people or objects in a local sequence. Their
records give the interval of the whole clip and express the order of
events within it in words. Five-minute clips trace events and transitions
across scenes, providing context for locating an earlier occurrence or
following an entity into a later scene.

\begin{table}[H]
\centering
\caption{\textbf{Information captured at the three temporal scales.}}
\label{tab:construction_information_scales}
\begingroup
\small
\setlength{\tabcolsep}{6pt}
\renewcommand{\arraystretch}{1.2}
\begin{tabularx}{\linewidth}{@{}>{\raggedright\arraybackslash}p{0.17\linewidth}>{\raggedright\arraybackslash}X@{}}
\toprule
\textbf{Temporal scale} & \textbf{Recorded information} \\
\midrule
5 seconds & Visible entities and objects, attributes, actions, spatial and interaction relations, visible text, and local state changes. \\
\midrule
1 minute & Major events, character and object state changes, and local temporal relationships. \\
\midrule
5 minutes & Broader event progression, scene transitions, longer-term states, and relationship changes. \\
\bottomrule
\end{tabularx}
\endgroup
\end{table}

For example, a five-second description from \texttt{film\_tv\_02\_001}
records a floral painting behind the speakers and subtitles at the bottom
of the frame. In the same video, the one-minute description for
5880--5940\,s follows a conversation from a proposed trip to the reason
for buying a piano and the response to that proposal. The five-minute
description for 5700--6000\,s places the conversation within a sequence
that moves from a street to a house and then to the living room. These
records give the next stage both specific details to examine and cues
for finding them in context. Each observation can be traced to its
source clip.

\paragraph{Step 2. Select candidate evidence for the target question type.}
For each of the 17 question subtypes, we specify what the respondent must
recover or infer and where the supporting evidence must occur
(Section~\ref{sec:evidence_dependency}). Codex examines the description
records against these requirements and proposes an observation or a set
of observations. For a Current question, the required information must
be visible at the time of the query. For a Historical question, an
earlier event, entity, or observation window supplies the target. A
recall item may require finding several past facts. A multi-evidence
reasoning item additionally needs a conclusion to which each selected
fact contributes a necessary premise.

The choice of intervals follows the intended dependency. A Localized
Current or Historical question draws on one interval, while a Distributed
one draws on separated intervals. For a Joint question, we first
identify what the current scene contributes and what the respondent
must recover from earlier footage. Its Localized or Distributed
designation describes whether that historical evidence lies in one
interval or several.

\paragraph{Step 3. First human review of the selected evidence.}
Reviewers first screen the selected candidates for suitability to the
intended task. They then inspect the relevant source footage to verify
the entity or event, the fact or relationship attributed to it, and its
supporting interval. When records at different scales disagree about
who performed an action, for example, the footage
determines which attribution can be used. Candidates with unresolved
contradictions or uncertain support are excluded. Only observations
that pass this review are used to construct questions and reference answers.

\Needspace{5\baselineskip}
\paragraph{Step 4. Construct the question and reference answer together.}
Using the verified evidence and the subtype definition, the Codex workflow
drafts an open-ended question and its reference answer. The wording must
identify the intended event or entity while leaving the requested fact
or relationship for the respondent to determine, as illustrated by
the fuel question in Table~\ref{tab:question_setting}.

The question record links the wording, canonical answer, and accepted
alternatives to the source video and supporting intervals. It also
records the subtype, evidence locality and distribution, and query time.
For delayed Historical questions, the wording and reference answer
continue to refer to the same event. The delays are measured from the
point when all supporting evidence has appeared and any observation
window specified by the question has ended
(Section~\ref{sec:evaluation_protocol}).

\paragraph{Step 5. Second human review of the question and reference answer.}
Reviewers check the completed question and reference
answer against the video. They must be able to identify what the
question asks and recover the reference answer from the designated
evidence. They then check whether the final wording preserves the
intended dependencies. For multi-evidence reasoning, removing any
designated supporting observation must leave the conclusion undetermined.
For Joint questions, neither the current scene nor the history alone
may determine the answer. In the sack example in
Table~\ref{tab:question_setting}, the current frame identifies
which sack the porter is lifting. Earlier footage supplies the identity
of its occupant. If the wording alone identified the sack, a respondent
could recover the answer from history without looking at the current frame.

For Historical questions, reviewers also examine subsequent footage.
If it repeats the answer-bearing observation or supplies a different
way to obtain the answer, the question is excluded. This check matters
for matched queries at increasing delays, since a later repetition
could make an earlier fact appear to have been retained. Questions
with reviewer disagreement are discarded. The final set contains 4,337 questions. Appendix~\ref{app:scorer_validation} reports the
separate audit of final questions, references, and model responses.

\clearpage
\FloatBarrier
\clearpage
\section{Evaluation Protocol and Scoring Reliability}
\label{app:evaluation_prompts}

\subsection{Streaming Evaluation and Query Scheduling}
\label{app:streaming_evaluation}
Frames arrive at 8 FPS. Each method samples from the frames available at that time using the settings in Appendix~\ref{app:evaluation_configurations}. The video clock continues during memory updates and answering. Current and Joint questions are released at their annotated times. For every question, the answerer can use only the frames and memory available when the question arrives.

\paragraph{Matched Historical questions.}
For a Historical question $q$, let $b_q$ be the time by which all supporting evidence has appeared and any observation window specified in the question has ended. We ask the question at $t_q=b_q+\delta$, with delays of 30 seconds and 1, 2, 4, 8, 16, and 32 minutes. Wording, supporting evidence, and reference answers are fixed across delays. Each delay is evaluated in a separate run starting with fresh memory, so an earlier repetition of the question cannot influence a later answer.

In selecting videos for this comparison, we excluded videos with too few eligible questions at 32 minutes and videos with markedly uneven numbers of eligible questions across delays. The first five delays share the same 1,217 questions. At 16 and 32 minutes, we retain a question from this set if its scheduled query falls within its source video ($b_q+\delta\leq T_v$, where $T_v$ is the video duration). All methods receive the same questions at a given delay. The first five scores therefore track the same evidence as the stream progresses. At the two longest delays, the composition of the question set also changes (Table~\ref{tab:historical_eligibility}).

\begin{table}[H]
\centering
\caption{\textbf{Historical questions evaluated at each delay.} The first row applies separately to each listed delay.}
\label{tab:historical_eligibility}
\small
\setlength{\tabcolsep}{8pt}
\renewcommand{\arraystretch}{1.15}
\begin{tabular}{@{}lr@{}}
\toprule
\textbf{Delay} & \textbf{Questions evaluated} \\
\midrule
30~s, 1, 2, 4, 8~min & 1,217 \\
16~min & 1,018 \\
32~min & 706 \\
\bottomrule
\end{tabular}
\end{table}

\subsection{Answer Scoring and Task-Specific Criteria}
Models are asked to answer in at most five words or return \texttt{REJECT} when evidence is insufficient. An exact match to the canonical answer or an accepted alternative receives full credit. For other substantive responses, Qwen3.5-9B receives the question, question type, answer type, model response, and reference answers as text. The judge runs in BF16 with thinking disabled, temperature zero, and a maximum of 32 output tokens. A complete, semantically equivalent answer earns 1. An answer earns 0.5 when it supplies an identifiable, correct part of the requested information and the task permits partial credit. Incorrect answers, empty outputs, refusals, and answer-generation failures (including out-of-memory errors) receive 0 and remain in the denominator. If the judge returns an invalid score, we retry once with a format instruction and refer unresolved cases for human scoring under the same criteria.

\paragraph{Current.}
We score the visible detail that the question asks for. Counts must be exact, although equivalent expressions such as \emph{two} and \emph{a pair} receive the same score. Spatial answers must preserve the requested relation. For object or attribute questions, a correct but less specific identification can earn half credit. For visible text, a minor omission can earn half credit if the response still identifies the requested text. When the requested detail is the step number, \texttt{STEP ONE} for \texttt{STEP THREE} receives zero because the number is wrong.

\paragraph{Historical.}
The answer must recover the requested past observation or state the conclusion supported by the designated observations. A causal question requires the cause, a comparison requires the comparative result, and a sequence requires the relevant events in order. A correct substantive part of a separable, multipart answer can earn half credit. Mentioning an event involved in a comparison earns zero if the response leaves the comparison unanswered.

\paragraph{Joint.}
For a question asking how a state changed, full credit requires both the earlier and current states in the correct direction. Both states in the correct direction with one imprecisely identified can earn half credit. Giving only one endpoint or reversing the change earns zero. A question that uses history to identify an entity in the current view can take the entity's name as a complete answer. For a question asking how a process evolved, the response must convey the requested stages in order.

\subsection{Score Aggregation}
For each method, we first average the individual ratings on the $\{0,0.5,1\}$ scale across the eligible questions of each subtype, pooling questions across videos. For a Historical subtype $c$ and delay $\delta$, let $Q_{c,\delta}$ denote its eligible questions and $s_{q,\delta}$ the rating for question $q$. The subtype score and the Historical family score at that delay are
\begin{equation}
S_{c,\delta}=\frac{100}{|Q_{c,\delta}|}\sum_{q\in Q_{c,\delta}}s_{q,\delta},
\qquad
H_\delta=\frac{1}{7}\sum_{c\in\mathcal H}S_{c,\delta},
\end{equation}
where $\mathcal H$ is the set of seven Historical subtypes. Current and Joint scores follow the same two steps at their annotated query times, with five subtypes each. The three \emph{Avg.} columns in Table~\ref{tab:main_results} therefore give equal weight to each subtype within its family.

The main Historical average is $(H_{30}+H_{60}+H_{120}+H_{240}+H_{480})/5$, with delays expressed in seconds. Each Historical subtype column in the main table likewise averages that subtype's scores over those five delays. Table~\ref{tab:history_results} reports the 16- and 32-minute scores separately on the smaller sets in Table~\ref{tab:historical_eligibility}.

\subsection{Scorer Calibration and Reliability}
\label{app:scorer_validation}
\paragraph{Calibrating the scoring prompts.}
We drew a development set at random in proportion to question-type frequencies and excluded it from the main results. We compared the judge's ratings with human ratings, examined disagreements against the source videos, and refined the instructions for the corresponding question types. The criteria distinguish wording differences that preserve the requested fact from errors that change it, as in the step-number example above. After refinement, the judge and human raters had no disagreements between 0 and 1 on the development set. Disagreements involving partial credit (0.5 versus 0 or 1) occurred in 3\% of cases. We then fixed the prompts for all methods.

\paragraph{Auditing questions and reference answers.}
We randomly sampled 300 model responses from the full evaluation output pool, spanning multiple videos. For each response, a reviewer who had not written the corresponding question checked the question, its reference answers, and the model response against the source video. The reviewer found no factual errors in the audited questions or reference answers. Two questions admitted valid answers absent from their references. In one case, the video confirmed that \emph{flowers} and \emph{a bouquet} identified the same item at the level of detail the question required. We added the valid alternatives and rechecked every evaluated method's score on both questions.

\paragraph{Checking agreement on final responses.}
For the same 300 responses, we compared the automated rating with the reviewer's rating based on the video. They disagreed on six responses (2\%). Summed over all 300 responses on the original $\{0,0.5,1\}$ scale, the automated and human totals differed by at most one point. The six disagreements count individual responses. The one-point bound applies to the difference between totals over the full sample, where differences in opposite directions can cancel. Results from rescoring the final outputs with the DeepSeek and GPT APIs appear in Appendix~\ref{app:cross_family_judges}.
\FloatBarrier
\clearpage
\begingroup
\raggedbottom
\section{AsyncMemFlow Implementation and Memory Examples}
\label{app:memory_comparison}

\subsection{Asynchronous Memory Construction}
AsyncMemFlow processes completed one-minute video blocks in temporal order. Each block is represented by 30 chronological frames sampled at 0.5 FPS. The description model records entities, visual attributes, actions, state changes, and relationships, together with the source interval. Description generation runs in the background using a model configured separately from the online answerer.

Each newly completed description triggers a Memory Map update. Given the previous Map and the new descriptions, the organizer incorporates observations into scenes, groups related scenes into events, and updates scene summaries and event headings as additional context becomes available. The timestamped descriptions remain the underlying records, including when they are awaiting organization. In the Qwen-based experiments, the organizer is fixed to local Qwen3-VL-8B-Instruct, regardless of the description or answer model. The cross-family configurations replace all three roles together, as detailed in Appendix~\ref{app:evaluation_configurations}.

\paragraph{Background scheduling.}
Description generation takes priority over Map organization. When a new description task begins, ongoing organization is suspended; the organizer uses the background GPU's available time between description tasks to continue updating the Map. This scheduling allocates spare computation to memory organization without making description generation wait for an unfinished Map update. Completed descriptions remain accessible throughout this process. Online answering reads the latest completed Map together with descriptions awaiting organization, without waiting for either background operation to finish.

\paragraph{Causal availability.}
A video block becomes eligible for description only after its final frame arrives. Its description becomes accessible when generation completes and remains available as an ungrouped record until incorporated into a completed Map. An updated Map becomes available only after organization finishes. For each question, all inputs are fixed at arrival: a description or Map update that completes later can affect subsequent questions, but not the current answer. Thus, the availability of new textual evidence depends on description completion, not on completion of its organization.

\paragraph{Memory representation and budget.}
Each description retains its source interval and text. Scene summaries and event headings organize these records while preserving access to their details; an example is provided in Section~\ref{app:memory_map_example}. In our experiments, the context budget follows the maximum input context length of the model in use, rather than a smaller fixed textual-memory limit. When this budget is reached, the organizer compresses the content associated with the earliest event in the active context. Original captions remain stored: captions omitted from the active context are not loaded into GPU memory, rather than deleted. This conditional compression was not triggered in the EOV-Bench runs, including the longest video.

\subsection{Routing and Answer Inputs}
\label{app:asyncmemflow_configuration}
The answer model first makes a text-only routing decision from the question. It selects the raw-frame branch when confident that recent visual observations are sufficient; other questions follow the default branch. Table~\ref{tab:asyncmemflow_answer_inputs} summarizes the inputs supplied to each branch.

\begin{table}[H]
\centering
\caption{\textbf{Inputs to the two online answering branches.}}
\label{tab:asyncmemflow_answer_inputs}
\begingroup
\small
\setlength{\tabcolsep}{6pt}
\renewcommand{\arraystretch}{1.2}
\begin{tabularx}{\linewidth}{@{}>{\raggedright\arraybackslash}p{0.15\linewidth}X@{}}
\toprule
\textbf{Branch} & \textbf{Answer-model inputs} \\
\midrule
Raw-frame & Question and the latest five causal frames sampled at 1 FPS, or all available sampled frames if fewer than five exist. \\
\midrule
Default & Question, the latest completed Memory Map, available ungrouped descriptions, and all frames sampled at 0.2 FPS after the end of available textual coverage and up to question arrival. \\
\bottomrule
\end{tabularx}
\endgroup
\end{table}

Inputs are fixed at question arrival, and all visual inputs have timestamps no later than that time. Background updates continue to prepare memory for subsequent questions.

\paragraph{Supplemental visual evidence.}
For the default branch, textual coverage extends to the end of the latest video block whose description is already available, including descriptions not yet incorporated into the Map. Frames after this boundary and up to question arrival are sampled at 0.2 FPS, with no separate frame-count cap or truncation window. These frames cover the current incomplete block and any preceding block still being described. In the measured configurations, description generation completes within one block interval; with one-minute blocks, the uncovered history therefore remains below two minutes. This operating range follows from the measured processing speed rather than an imposed input limit. Map organization does not extend the uncovered interval, because completed descriptions remain accessible while it proceeds.

\paragraph{Routing accuracy.}
Routing accuracy is evaluated over all questions as the proportion correctly assigned to the Current or default branch, with Historical and Joint questions assigned to the default branch. The accuracies are 81.41\%, 91.57\%, and 96.61\% for the 2B, 4B, and 8B answerers, respectively.

\subsection{Memory Map Example}
\label{app:memory_map_example}
The following example presents a memory organization result for \emph{Mount Joy} (video \texttt{film\_tv/02/002}), using five descriptions spanning 4020--4320\,s. It illustrates the flat and organized representations rather than a memory snapshot available at a specified real-time query.

The five descriptions, denoted D1--D5 below, appear in both the flat and organized inputs. Their full text is listed once after the structural comparison. The event heading and scene summaries are translated from Chinese into English for presentation. Both description generation and memory organization use Qwen3-VL-8B-Instruct in this example. This excerpt displays records from newest to oldest for presentation; the experimental answer inputs order descriptions from earliest to latest in both Map-on and Map-off settings.

\begin{table}[H]
\centering
\caption{\textbf{Flat description order in the illustrated memory excerpt.}}
\label{tab:flat_memory_example}
\begingroup
\small
\setlength{\tabcolsep}{6pt}
\renewcommand{\arraystretch}{1.2}
\begin{tabularx}{\linewidth}{@{}l*{5}{>{\centering\arraybackslash}X}@{}}
\toprule
\textbf{Record} & \textbf{D1} & \textbf{D2} & \textbf{D3} & \textbf{D4} & \textbf{D5} \\
\midrule
Source interval (s) & 4260--4320 & 4200--4260 & 4140--4200 & 4080--4140 & 4020--4080 \\
\bottomrule
\end{tabularx}
\endgroup
\end{table}

\begin{table}[H]
\centering
\caption{\textbf{Memory Map organization of the same descriptions.}}
\label{tab:organized_memory_example}
\begingroup
\small
\setlength{\tabcolsep}{6pt}
\renewcommand{\arraystretch}{1.2}
\begin{tabularx}{\linewidth}{@{}>{\raggedright\arraybackslash}p{0.19\linewidth}>{\raggedright\arraybackslash}X@{}}
\toprule
\textbf{Source interval (s)} & \textbf{Map entry and associated descriptions} \\
\midrule
4260--4320 & \textbf{Ungrouped record:} D1. \\
\midrule
4200--4260 & \textbf{Completed scene within an ongoing event.} The woman lights a cigarette and places the burning cigarette on dry straw, which begins to burn; an orange glow subsequently appears in an upstairs window of the barn. \textbf{Description:} D2. \\
\midrule
3180--4200 & \textbf{Completed event.} Following and confrontation, relationship dynamics, and the barn gathering. \\
\cmidrule(lr){1-2}
\quad 4140--4200 & \textbf{Scene.} The woman is alone inside the barn; several men have a brief confrontation outside; the segment ends with close-ups of jeans being unbuttoned and an object being taken from a pocket. \textbf{Description:} D3. \\
\cmidrule(lr){1-2}
\quad 4020--4140 & \textbf{Scene.} The red-haired woman leads the man in green into a barn filled with cardboard boxes and shows him a black-and-white photograph of two people; after they kiss, the woman becomes uneasy and the man leaves. \textbf{Descriptions:} D4, D5. \\
\bottomrule
\end{tabularx}
\endgroup
\end{table}

For the completed event spanning 3180--4200\,s, the table displays its scenes from 4020\,s onward. The descriptions below present the visual observations, clip-relative timestamps, and uncertainty.

\label{app:shared_memory_descriptions}
\noindent\begin{minipage}{\linewidth}
\textbf{D1\quad 4260--4320\,s}\par\smallskip
At 0s, a solitary wooden barn is visible at night, one window faintly lit; by 7s, the frame cuts to near-blackness with blurred limbs moving. From 12s--20s, a woman turns her head, then a bearded man lies on leafy ground, blinking slowly into awareness. At 21s--26s, his face fills frame, eyes wide, mouth slightly open; another person's face is blurred behind him. At 29s, a man in yellow shirt and glasses appears startled beside a tank-top-clad figure; they assist or restrain someone. By 34s, figures flee into darkness; trees and distant lights appear. At 38s, the barn erupts in massive flames across its upper structure, fire climbing vertically, embers falling. The blaze intensifies through 55s, obscuring details. At 58s--60s, a backlit figure runs leftward, away from the fully engulfed building.
\end{minipage}\par\medskip

\noindent\begin{minipage}{\linewidth}
\textbf{D2\quad 4200--4260\,s}\par\smallskip
The video begins in near-darkness with a blurry view of fabric, then reveals an orange pill bottle lying among dry straw. A hand enters frame, lighting a cigarette with a metallic lighter; the woman's face is partially visible, eyes lowered. She places the lit cigarette onto the straw, which quickly ignites into a bright, flickering flame. After \textasciitilde{}10 seconds of burning straw, the scene cuts to a static wide shot of a wooden barn against a dusky sky; clouds drift slowly as light fades. Near the end, a small window on the barn's upper gable briefly emits an orange glow before going dark again.
\end{minipage}\par\medskip

\noindent\begin{minipage}{\linewidth}
\textbf{D3\quad 4140--4200\,s}\par\smallskip
A woman with short brown hair and bangs, wearing a black tank top and necklace, looks upward with a faint smile in a dim interior (0:00--0:04). A man in a green t-shirt walks past stacks of labeled cardboard boxes, turns to face her, and they exchange a silent glance before he walks away (0:05--0:19). She remains standing, holding a red plastic cup, glancing down (0:20--0:25). Cut to exterior night: the man walks toward a house, meets another man, and they briefly grapple or gesture aggressively (0:26--0:33); two others join them. After a black screen (0:34--0:36), the woman is again shown indoors, still holding the cup, then suddenly leans forward and exits frame (0:37--0:43). Blurry motion follows, then tight close-ups show hands unbuttoning denim jeans and pulling something from the front pocket (0:51--0:56).
\end{minipage}\par\medskip

\noindent\begin{minipage}{\linewidth}
\textbf{D4\quad 4080--4140\,s}\par\smallskip
A woman holds up a black-and-white photo of two men and shows it to a man beside her; they lean in together, examining it with focused expressions. She speaks, he responds, and their proximity increases until they kiss intimately for several seconds. Immediately after breaking the kiss, the woman's expression shifts to alarm or distress as she speaks again; the man reacts with furrowed brow and looks away. He then stands and walks toward the background where cardboard boxes are stacked, while she remains seated, lowering her head with a troubled look. The lighting remains low and warm throughout, with shallow depth of field keeping background elements blurred.
\end{minipage}\par\medskip

\noindent\begin{minipage}{\linewidth}
\textbf{D5\quad 4020--4080\,s}\par\smallskip
At 0:00, a woman with red hair and heavy eye makeup looks off-camera with a slight smile in a nighttime outdoor scene with blurred multicolored lights. By 0:03, she stands, holding a red cup, and moves past a seated man in a green shirt near a drained concrete skate bowl; others are present, one crouching, another holding a skateboard. At 0:08, the man rises and follows her as she walks toward a large wooden structure. From 0:13--0:22, they walk side-by-side across a grassy/dirt area toward the barn's double doors. At 0:23, they enter a dim interior filled with stacked cardboard boxes; she gestures rightward while speaking, still holding the cup, and he follows closely. Between 0:27--0:33, they navigate narrow aisles between boxes; camera pans slightly to keep them centered. From 0:34 onward, they pause intermittently---she turns to him, smiles, talks, and gestures again; he listens, nods, and at 0:51 bends down near a pile of boxes while she watches. She continues walking forward, glancing back, until the clip ends at 0:59 with her looking toward the camera, smiling, holding a small white object (possibly a photo or card) in her left hand.
\end{minipage}\par\medskip

\clearpage
\endgroup
\FloatBarrier
\clearpage
\clearpage
\makeatletter
\setlength{\@fptop}{0pt}
\setlength{\@fpsep}{12pt}
\makeatother
\section{Experimental Configurations}
\label{app:evaluation_configurations}

\subsection{Shared Experimental Settings}
The main latency measurements use a shared NVIDIA A100 80GB GPU for local background computation and online answering. The background scheduling and query-time inputs of AsyncMemFlow are described in Appendix~\ref{app:memory_comparison}.

The main AsyncMemFlow configuration uses separate local Qwen3-VL-8B-Instruct models for video description and answering. Qwen3-VL-8B-Instruct also organizes the Memory Map, including in runs with Qwen-VL-Max descriptions.

Online response time is measured from question arrival to completion of the answer while background processing continues. In Table~\ref{tab:description_compute}, memory-update time denotes mean node-generation time for OASIS and mean time per video block to complete a Memory Map update for AsyncMemFlow. Both quantities are measured in our evaluation using the same model configurations as the corresponding accuracy results.

\subsection{Baseline Configurations}
Table~\ref{tab:baseline_configurations} lists the model checkpoints, sampling rates, and memory settings used in the main experiments. HERMES (Q) and Flash-VStream (Q) use Qwen3-VL-8B-Instruct through training-free backbone substitution. StreamForest, VST, and Dispider use their released trained checkpoints.

\begin{table}[!htbp]
\centering
\caption{\textbf{Model checkpoints and inference configurations.}}
\label{tab:baseline_configurations}
\begingroup
\fontsize{7}{8.5}\selectfont
\setlength{\tabcolsep}{3pt}
\renewcommand{\arraystretch}{1.15}
\begin{tabularx}{\linewidth}{@{}>{\raggedright\arraybackslash}p{0.14\linewidth}>{\raggedright\arraybackslash}p{0.22\linewidth}>{\centering\arraybackslash}p{0.07\linewidth}>{\raggedright\arraybackslash}X@{}}
\toprule
\textbf{Method} & \textbf{Model checkpoints} & \textbf{FPS} & \textbf{Memory and inference configuration} \\
\midrule
Raw Frames & Qwen3-VL-8B-Instruct~\citep{bai2025qwen3vltechnicalreport} / LLaVA-OneVision-Qwen2-7B & 1 & Latest five frames. \\
\midrule
ReKV~\citep{ICLR2025_67a9b444} & LLaVA-OneVision-Qwen2-7B & 1 & Local KV budget: 15,000; retrieval top-$k$: 64; retrieval chunk size: 1; encoding chunk size: 8 frames. \\
\midrule
HERMES (L)~\citep{zhang-etal-2026-hermes} & LLaVA-OneVision-Qwen2-7B & 0.5 & KV budget: 6,000; encoding chunk size: 16 frames; QA write-back disabled for Historical and Joint queries. \\
\midrule
HERMES (Q)~\citep{zhang-etal-2026-hermes} & Qwen3-VL-8B-Instruct & 0.5 & KV budget: 6,000; encoding chunk size: 16 frames; maximum image pixels: 262,144; QA write-back disabled. \\
\midrule
StreamForest~\citep{NEURIPS2025_6dd91fec} & StreamForest-Qwen2-7B & 1 & Short-term window sizes: [1, 18], with token allocations [729, 128]; long-term budget: 5,120 tokens, with 64 tokens per frame; event-splitting window: 8. \\
\midrule
Flash-VStream (Q)~\citep{zhang2025flashvstream} & Qwen3-VL-8B-Instruct & 1 & Incremental temporal compression and spatial-detail selection; temporal/spatial length parameters: 120/60, corresponding to 60/30 two-frame units. \\
\midrule
VST~\citep{10.1007/978-3-032-37255-0_5} & VST-7B (Qwen2.5-VL-based) & 1 & Visual chunk budget: 8,192 tokens; FIFO textual memory: 12,000 characters; separate, identical frozen model instances for memory generation and answering. \\
\midrule
Dispider~\citep{Qian_2025} & Qwen2-1.5B Perception--Decision + Qwen2-7B Reaction & 1 & 16-frame clips; cached completed-clip representations; native decision, retrieval, and reaction components; \texttt{SILENT} retained for non-triggered responses. \\
\midrule
OASIS~\citep{liang2026oasis} & Qwen3-VL-8B-Instruct + Qwen3-Embedding-0.6B & 1 & Current-window capacity: 16 frames; short-term buffer: 32 frames; event-node capacity: 16 frames; 256 tokens per frame; maximum 4 root nodes; up to 2 retrieved events. \\
\bottomrule
\end{tabularx}
\endgroup
\end{table}

\FloatBarrier
\paragraph{Cross-family adaptation.}
For cross-family evaluation, we replace each method's adaptable multimodal components while retaining its sampling rate, memory budget, and memory strategy. Both OASIS variants retain Qwen3-Embedding-0.6B and use OASIS-generated descriptions and node summaries. The direct-access variant changes how these records are supplied to the answerer. Appendix~\ref{app:cross_family_models} specifies the backbone used for each AsyncMemFlow role in the full-backbone and describer-only experiments.

\paragraph{Evaluation interface and failures.}
All methods receive questions at the scheduled times and use information available at question arrival. Failed, empty, rejected, and out-of-memory responses score zero and remain in the denominator.

\subsection{Ablation and Memory-Access Configurations}

\paragraph{Description window and model.}
Table~\ref{tab:description_compute} uses an 8B answer model with the Memory Map enabled. The local Qwen3-VL-8B-Instruct describer processes 30-second or 60-second blocks, sampled at 0.5 FPS to produce 15 or 30 frames per block. The Qwen-VL-Max configuration uses 60-second blocks with 30 frames. The answer model and Map settings remain fixed across these configurations.

\paragraph{Memory organization.}
The Map-on/off pairs in Table~\ref{tab:ablation} compare the two memory representations under the corresponding description-model and answer-model configurations, using the same questions at each delay. Both settings present descriptions in chronological order, from earliest to latest. Map-off supplies the available original video descriptions directly to the answerer. Map-on places descriptions within the Memory Map structure, which additionally provides scene summaries, event headings, and hierarchical grouping. The comparison therefore evaluates the contribution of this organization and its associated summaries. Information is fixed at question arrival, following the availability rules in Appendix~\ref{app:memory_comparison}.

\paragraph{Memory retrieval.}
Table~\ref{tab:memory_access} evaluates BM25~\citep{robertson2009bm25}, PPR~\citep{jeh2003personalized}, Word TF-IDF~\citep{salton1988termweighting}, and Qwen-guided retrieval over one-minute description records, with top-$k$ fixed to five. Qwen-guided retrieval uses Qwen3-VL-8B-Instruct to select records in a single call, following instructions adapted from OASIS~\citep{liang2026oasis}. The selected original descriptions are supplied to the answerer without scene summaries or event headings, which organize related descriptions and can span multiple video blocks. The underlying stored memory and answer model remain fixed within each system. Direct memory access supplies the complete structured memory, including its summaries.

Recall@5 measures complete evidence coverage at the question level. An annotated evidence item is matched when a retrieved description corresponds to the one-minute video block containing it. A question receives a retrieval score of one only if the five selected records collectively cover all its required supporting evidence, and zero otherwise. Recall@5 is the percentage of evaluated questions satisfying this condition; retrieving only part of a question's required evidence receives no retrieval credit. Historical answer scores are computed separately using the shared scoring protocol.

\paragraph{OASIS memory access.}
\label{app:oasis_our_version}
The OASIS variant in Table~\ref{tab:history_results} places OASIS-generated descriptions and node summaries, including root summaries, into the structured memory container used by our direct-access interface. All textual content is produced by OASIS; this adaptation changes how the existing content is assembled and supplied to the answerer. The container is serialized into a single textual context and provided directly together with recent visual observations, with query-dependent retrieval disabled. Table~\ref{tab:history_results} uses the original retrieval configuration for the OASIS baseline. Table~\ref{tab:memory_access} changes its retrieval count to top-5 while keeping the remaining settings unchanged. Both tables use the same direct-access variant.

\subsection{Evaluation on Other Benchmarks}
On OVO-Bench~\citep{11094230} and StreamArena~\citep{zhang2026streamarenacontinuousinteractivelonghorizon}, AsyncMemFlow uses separate Qwen3-VL-8B-Instruct models for description and answering. We follow each benchmark's official input and scoring protocols. For OVO-Bench, we average task scores within each group (six tasks for Real-Time Visual Perception, three for Backward Tracing, and three for Forward Active Responding). For StreamArena, we report real-time perception and historical retrospection at the four official delay levels (L1--L4). The baseline scores used for comparison come from the StreamArena paper~\citep{zhang2026streamarenacontinuousinteractivelonghorizon}.

\FloatBarrier
\FloatBarrier
\section{Broader Ablations}
\label{app:broader_ablations}

\subsection{Cross-family Model Evaluations}
\label{app:cross_family_models}

Qwen3.5-9B scores all configurations. Table~\ref{tab:cross_family_results} compares OASIS, HERMES, and AsyncMemFlow across three backbones. In these AsyncMemFlow runs, the same backbone describes the video and answers questions. Table~\ref{tab:cross_family_describers} compares historical retention across seven delays. For each AsyncMemFlow pair, Map off supplies descriptions in chronological order, and Map on uses the answering backbone to organize them into the Memory Map. Appendix~\ref{app:evaluation_configurations} gives the adaptation details.

\begin{table}[H]
\centering
\caption{Method comparison by backbone on EOV-Bench. Current, Historical, and Joint use the scoring protocol of Table~\ref{tab:main_results}. Historical averages the five matched delays from 30 seconds to eight minutes.}
\label{tab:cross_family_results}
\footnotesize
\setlength{\tabcolsep}{3pt}
\begin{tabular*}{\linewidth}{@{}l@{\extracolsep{\fill}}rrrrrrrrr@{}}
\toprule
& \multicolumn{3}{c}{\textbf{Qwen3-VL-8B-Instruct}} & \multicolumn{3}{c}{\textbf{InternVL2.5-8B}} & \multicolumn{3}{c}{\textbf{GLM-4.6V-Flash}} \\
\cmidrule(lr){2-4}\cmidrule(lr){5-7}\cmidrule(l){8-10}
\textbf{Method} & Current & Historical & Joint & Current & Historical & Joint & Current & Historical & Joint \\
\midrule
OASIS & 71.29 & 28.88 & 35.95 & 64.33 & 23.27 & 31.51 & 70.03 & 28.98 & 33.82 \\
HERMES & 66.50 & 33.60 & 21.36 & 62.79 & 24.87 & 19.56 & 65.57 & 25.53 & 20.85 \\
\midrule
AsyncMemFlow (Map off) & 75.19 & 32.90 & 33.24 & 69.25 & 27.30 & 30.16 & 73.19 & 30.19 & 31.77 \\
AsyncMemFlow (Map on) & 75.19 & 43.93 & 43.78 & 69.25 & 37.25 & 37.83 & 73.19 & 41.27 & 39.39 \\
\bottomrule
\end{tabular*}
\par\smallskip
\begin{minipage}{\linewidth}
\end{minipage}
\end{table}

With InternVL2.5-8B, our direct memory access raises OASIS's five-delay mean Historical score from 23.27 to 30.14. With GLM-4.6V-Flash, it rises from 28.98 to 33.46. Within each pair, the backbone, descriptions, and node summaries are held fixed. The gains across both backbones therefore support the benefit of supplying the stored memory directly to the answerer.

\begin{table}[H]
\centering
\caption{AsyncMemFlow Historical scores across seven query delays. Each Map-off/on pair uses the same describer and answerer. At each delay, all configurations use the same eligible questions (Table~\ref{tab:historical_eligibility}).}
\label{tab:cross_family_describers}
\small
\setlength{\tabcolsep}{4pt}
\begin{tabular*}{\linewidth}{@{}l@{\extracolsep{\fill}}llrrrrrrr@{}}
\toprule
\textbf{Describer} & \textbf{Answerer} & \textbf{Map} & \textbf{30\,s} & \textbf{1\,min} & \textbf{2\,min} & \textbf{4\,min} & \textbf{8\,min} & \textbf{16\,min} & \textbf{32\,min} \\
\midrule
GLM & Qwen & Off & 42.79 & 32.52 & 26.90 & 23.95 & 24.90 & 24.41 & 24.13 \\
GLM & Qwen & On & 47.42 & 43.12 & 38.93 & 40.48 & 39.26 & 41.56 & 41.11 \\
\addlinespace[2pt]
InternVL & Qwen & Off & 39.47 & 31.32 & 26.70 & 25.10 & 24.89 & 24.66 & 24.05 \\
InternVL & Qwen & On & 48.77 & 41.43 & 38.24 & 39.53 & 35.42 & 37.86 & 39.22 \\
\addlinespace[2pt]
InternVL & InternVL & Off & 33.05 & 28.22 & 26.24 & 24.68 & 24.29 & 24.01 & 23.21 \\
InternVL & InternVL & On & 42.21 & 40.56 & 32.84 & 35.25 & 35.40 & 36.45 & 35.55 \\
\addlinespace[2pt]
GLM & GLM & Off & 38.88 & 30.15 & 26.49 & 28.23 & 27.20 & 26.44 & 26.18 \\
GLM & GLM & On & 48.19 & 41.79 & 39.17 & 39.85 & 37.35 & 37.03 & 37.50 \\
\bottomrule
\end{tabular*}
\par\smallskip
\footnotesize
Qwen, GLM, and InternVL denote Qwen3-VL-8B-Instruct, GLM-4.6V-Flash, and InternVL2.5-8B, respectively.
\end{table}

\subsection{Cross-family Judge Evaluations}
\label{app:cross_family_judges}

We rescore the final answers from every method in Table~\ref{tab:main_results} with DeepSeek-V4.1 and GPT 5.6 Luna. All three judges use the same task-specific scoring criteria, reference answers, and accepted alternatives. Table~\ref{tab:cross_judge_results} compares their scores on the same questions and responses. The key relative rankings among the methods in Table~\ref{tab:main_results} remain consistent across judges. Appendix~\ref{app:scorer_validation} describes calibration and the independent human audit.

\begin{table}[H]
\centering
\caption{Judge comparison on EOV-Bench. Each cell reports Qwen3.5-9B / DeepSeek-V4.1 / GPT 5.6 Luna scores, in that order. The first score reproduces the corresponding average in Table~\ref{tab:main_results}.}
\label{tab:cross_judge_results}
\small
\begin{tabular*}{\linewidth}{@{}l@{\extracolsep{\fill}}ccc@{}}
\toprule
\textbf{Method} & \textbf{Current} & \textbf{Historical} & \textbf{Joint} \\
\midrule
Raw Frames (Qwen3-VL-8B) & 75.44 / 78.28 / 77.31 & 5.36 / 7.84 / 5.99 & 7.64 / 5.30 / 6.34 \\
Raw Frames (LLaVA-OneVision) & 62.32 / 63.14 / 63.34 & 6.99 / 7.31 / 7.47 & 8.03 / 8.44 / 8.57 \\
ReKV & 35.39 / 36.24 / 35.54 & 17.02 / 17.77 / 17.98 & 7.38\textsuperscript{*} / 7.51 / 7.03 \\
HERMES (Q) & 66.50 / 66.30 / 67.04 & 33.60 / 33.71 / 33.79 & 21.36 / 21.48 / 21.67 \\
HERMES (L) & 56.27 / 56.99 / 56.75 & 16.95 / 17.10 / 16.81 & 13.76 / 14.14 / 13.80 \\
Flash-VStream (Q) & 54.18 / 54.33 / 54.24 & 14.73 / 15.38 / 14.54 & 4.87 / 5.75 / 5.37 \\
OASIS & 71.29 / 73.51 / 71.95 & 28.88 / 29.07 / 29.14 & 35.95 / 36.35 / 36.22 \\
StreamForest & 56.13 / 57.64 / 57.12 & 25.26 / 25.35 / 25.82 & 12.24 / 13.08 / 12.69 \\
VST & 62.03\textsuperscript{*} / 62.11 / 62.79 & 27.91\textsuperscript{*} / 28.58 / 28.22 & 17.19\textsuperscript{*} / 17.62 / 17.54 \\
Dispider & 38.23\textsuperscript{*} / 38.34 / 38.99 & 22.33 / 22.46 / 22.99 & 12.04\textsuperscript{*} / 12.50 / 12.43 \\
AsyncMemFlow (local 8B) & 75.19 / 78.03 / 77.06 & 43.93 / 44.80 / 44.83 & 43.78 / 43.71 / 44.57 \\
\bottomrule
\end{tabular*}
\par\smallskip
\begin{minipage}{\linewidth}
\footnotesize
Scores are on a 0--100 scale and weight question subtypes equally within each column. Historical scores average the five delays from 30 seconds to eight minutes on 1,217 shared questions. (Q) and (L) denote Qwen3-VL-8B and LLaVA-OneVision. \textsuperscript{*}Runtime failures, including out-of-memory errors, count as zero.
\end{minipage}
\end{table}

\end{document}